\documentclass[10pt,twocolumn,letterpaper]{article}

\usepackage{wacv}

\usepackage{graphicx}
\usepackage{amsmath}
\usepackage{amssymb}
\usepackage{booktabs}
\usepackage{xcolor,colortbl}
\usepackage{nicefrac, xfrac}

\usepackage{gensymb}

\usepackage{algorithm}
\usepackage{algpseudocode}
\usepackage{titling}

\usepackage{tikz}
\usetikzlibrary{calc}

\usetikzlibrary{shadings}

\usepackage[pagebackref,breaklinks,colorlinks]{hyperref}

\usepackage[capitalize]{cleveref}
\crefname{section}{Sec.}{Secs.}
\Crefname{section}{Section}{Sections}
\Crefname{table}{Table}{Tables}
\crefname{table}{Tab.}{Tabs.}

\def\wacvPaperID{1410} 
\def\confName{WACV}
\def\confYear{2025}

\begin{document}

\title{RayGauss: Volumetric Gaussian-Based Ray Casting \\for Photorealistic Novel View Synthesis}

\author{Hugo Blanc, Jean-Emmanuel Deschaud, Alexis Paljic\\
Centre for Robotics, Mines Paris - PSL, PSL University, 75006 Paris, France\\
{\tt\small \{hugo.blanc,jean-emmanuel.deschaud,alexis.paljic\}@minesparis.psl.eu}
}
\date{}

\maketitle

\begin{abstract}

Differentiable volumetric rendering-based methods made significant progress in novel view synthesis. On one hand, innovative methods have replaced the Neural Radiance Fields (NeRF) network with locally parameterized structures, enabling high-quality renderings in a reasonable time. On the other hand, approaches have used differentiable splatting instead of NeRF's ray casting to optimize radiance fields rapidly using Gaussian kernels, allowing for fine adaptation to the scene.
However, differentiable ray casting of irregularly spaced kernels has been scarcely explored, while splatting, despite enabling fast rendering times, is susceptible to clearly visible artifacts.\\
Our work closes this gap by providing a physically consistent formulation of the emitted radiance \( c \) and density \( \sigma \), decomposed with Gaussian functions associated with Spherical Gaussians/Harmonics for all-frequency colorimetric representation. We also introduce a method enabling differentiable ray casting of irregularly distributed Gaussians using an algorithm that integrates radiance fields slab by slab and leverages a BVH structure. This allows our approach to finely adapt to the scene while avoiding splatting artifacts. As a result, we achieve superior rendering quality compared to the state-of-the-art while maintaining reasonable training times and achieving inference speeds of 25 FPS on the Blender dataset. The associated code is available at: \href{https://github.com/hugobl1/ray_gauss}{github.com/hugobl1/ray\_gauss}.

\end{abstract}

\begin{figure}[ht]
  \centering
   \includegraphics[trim={0cm 0cm 0cm 0cm},clip,width=\columnwidth]{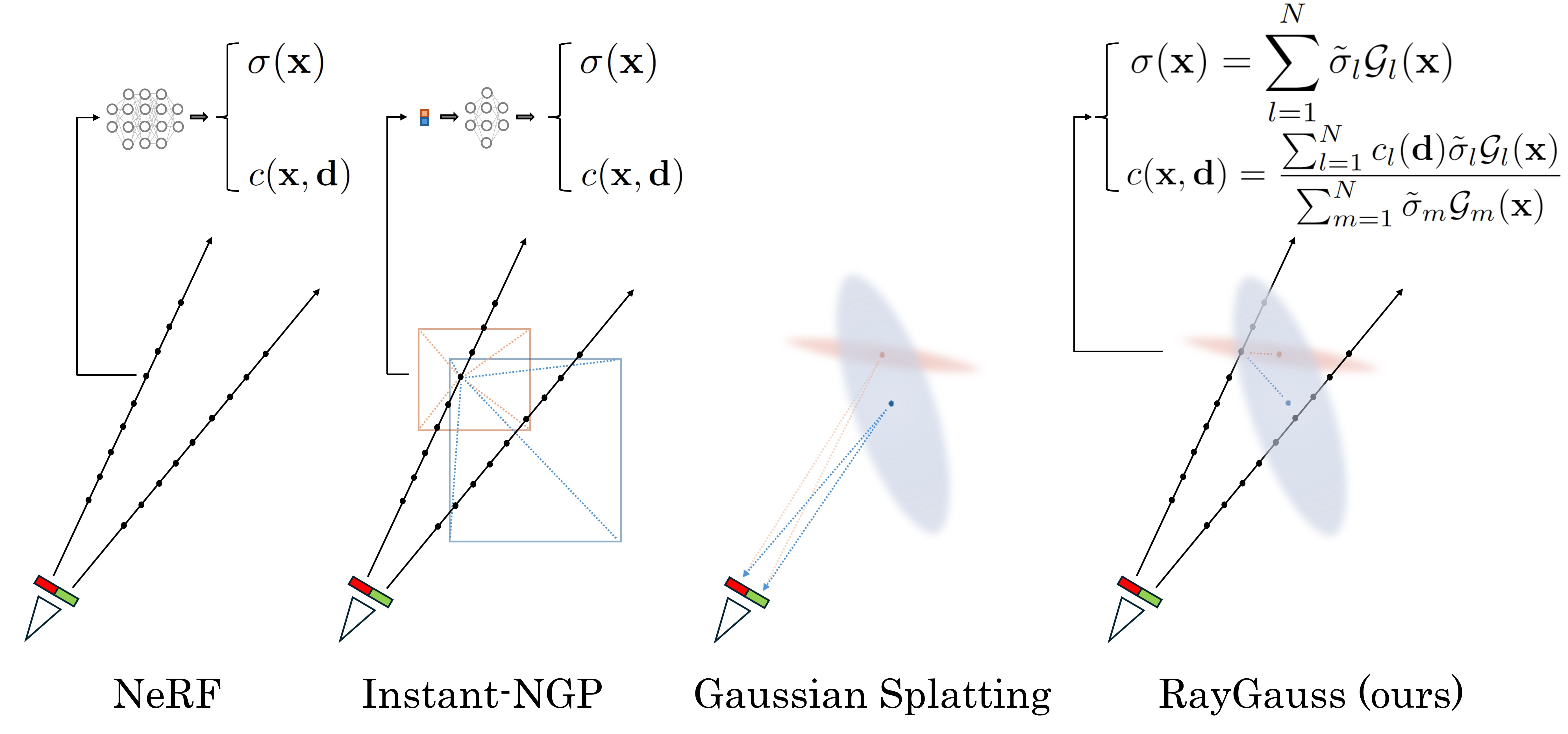}
   \caption{Our method (RayGauss) compared to the main radiance field rendering methods}
   \label{fig:figure2}
\end{figure}

\section{Introduction}
\label{sec:intro}
Novel View Synthesis (NVS), which combines existing views of a scene to generate images from unknown viewpoints, saw significant advancements with the publication of Neural Radiance Fields (NeRF) \cite{Neural_Radiance_Fields}. This novel approach leverages a differentiable physical rendering algorithm \cite{MaxOpt}, to learn radiance fields through a neural network, enabling the generation of high-quality photorealistic images. Since then, several works aimed at improving NeRF's performance, some focusing on finding more efficient representations of radiance fields in terms of computation time and rendering quality. 
Various data structures were investigated to substitute the neural network and represent the radiance fields: voxel grids \cite{Plenoxels}\cite{Direct_Voxel_Grid_optimization}\cite{Instant_NGP}, sets of tetrahedra \cite{Tetra_NeRF}, point clouds\cite{Point_NeRF}\cite{3D_Gaussian_Splatting}, hybrid representations\cite{chen2023neurbf}. These structures have the advantage of enabling the representation of radiance fields using local parameters, thereby reducing computation times by limiting field evaluation to local parameters instead of relying on a neural network representing the entire scene.
These methods also aim to enhance rendering quality by allowing a more precise representation of radiance fields through data structures better suited to the scene's geometry.

Our work follows this approach by proposing a novel definition of radiance fields based on their decomposition into a weighted sum of locally supported elliptical basis functions with optimizable positions, orientations, and scales. Unlike voxel grid-based methods, this allows for finer adaptation to the scene's geometry without limiting representation resolution. Furthermore, a combination of spherical harmonics (SH) and spherical Gaussians (SG) is associated with these basis functions to better represent high-frequency color variations such as specular reflections.
Additionally, we use volume ray casting, which produces high-quality images with fewer simplifications than splatting-based methods like 3D Gaussian Splatting\cite{3D_Gaussian_Splatting}. Indeed, splatting algorithms simplify the scene using multiple approximations, allowing faster rendering but causing many artifacts when deviating from the training views, such as flickering due to a sudden change in primitives ordering\cite{stopthepop}. For these reasons, our approach achieves higher rendering quality by combining the best of both worlds: Firstly, radiance fields are represented by elliptical functions associated with efficient SH/SG radiance parameters that adapt very well to the scene geometry and appearance. Secondly, the Volume Raycasting algorithm avoids the artifacts of Splatting, such as flickering. 

Furthermore, implementing the Raycasting algorithm with sparse primitives is challenging because evaluating the radiance fields at a given point requires knowing which primitives/basis functions contribute to the calculations, which is much less straightforward than when using a voxel grid, for example. Therefore, we introduce a ray casting algorithm for sparse basis functions that sequentially integrates the radiance field slab-by-slab, meaning it accumulates colorimetric properties along the ray in slab of space corresponding to multiple samples, rather than processing individual samples\cite{Knoll2019}. The implementation of this algorithm and its backpropagation relies on a Bounding Volume Hierarchy (BVH) implemented on GPU with the OptiX library. This allows interactive rendering times and reasonable training times.  Thus, our contributions are as follows:

\begin{itemize}
    \item Volume Ray Casting of radiance fields decomposed into elliptical basis functions combined with SH/SG for efficient adaptation to geometry and appearance.
    \item Efficient implementation of Volume Ray Casting and its backpropagation using a slab-by-slab integration algorithm along the ray and leveraging a Bounding Volume Hierarchy for fast ray-ellipsoïd intersection.
\end{itemize}

\section{Background and Related Work}
\label{sec:relatedwork}
We will introduce the theoretical basis of radiative transfer due to its essential role in our method and modern NVS approaches. Next, we will discuss the NeRF method and subsequent research on scene representation. Finally, we will discuss 3D Gaussian Splatting~\cite{3D_Gaussian_Splatting} and its flexible scene representation using Gaussians while highlighting the weaknesses of the splatting algorithm.


\textbf{Radiative Transfer and Volume Rendering:} Recent approaches to Novel View Synthesis, such as NeRF~\cite{Neural_Radiance_Fields}, Instant-NGP~\cite{Instant_NGP}, and 3D Gaussian Splatting~\cite{3D_Gaussian_Splatting}, leverage a differentiable volume rendering equation to optimize scene representation from images (Fig. \ref{fig:figure2}). Their rendering equations are derived from the radiative transfer equation in an absorbing and emitting medium, which models the variation in radiance as it travels through an infinitesimal volume element at position \(\mathbf{x}\) in direction $\omega$ \cite{MaxOpt}\cite{ewa}:
\begin{equation}
(\omega \cdot \nabla)L(\mathbf{x}, \omega) = -\sigma(\mathbf{x}) L(\mathbf{x}, \omega) + \sigma(\mathbf{x}) c(\mathbf{x}, \omega)
\label{eq:rad_transfer}
\end{equation}
where \( L(\mathbf{x}, \omega) \) is the radiance at position \( \mathbf{x} \) in direction \( \omega \),  \( \sigma(\mathbf{x}) \) the absorption coefficient and \( c(\mathbf{x}, \omega) \) the emitted radiance\cite{chandrasekhar2013radiative}. The first term quantifies the absorption of radiance, and the second term its emission. The solution to this non-homogeneous linear differential equation is:
\begin{equation}
    L(\mathbf{x}, \omega) = \int_{0}^{\infty}  c(\mathbf{y}, \omega)\sigma(\mathbf{y}) T(\mathbf{x}, \mathbf{y}) \, d\mathbf{y}
\end{equation}
by integrating along the ray reaching \(\mathbf{x}\) in the direction \( \omega \) defined by the points \({\mathbf{y}(t)} = \mathbf{x} - t \omega \),  and where \(T(\mathbf{x}, \mathbf{y}) = e^{-\int_{0}^{t} \sigma(\mathbf{x} - s\omega) \, ds} \)  is the transmittance between \(\mathbf{x}\) and \(\mathbf{y}\). 
This expression appears in the literature of Volumetric Light Transport Simulation \cite{novak18monte}. In the context of volumetric rendering, a modified parameterization is used: fictive rays denoted by \( \mathbf{r}(t) = \mathbf{o} + t \mathbf{d} \) originate from the camera center  \(\mathbf{o}\) and traverse the scene opposite to the light direction, assimilating \(\mathbf{d}\) to \(-\omega\). In computer vision, the focus is on the ray's color, interpreting radiance \(L\) as the color \(C\) of the ray, and the absorption coefficient \(\sigma\) renamed density. This parameterization leads to the classic volumetric rendering equation \cite{Neural_Radiance_Fields}\cite{ewa}:

\begin{equation}
\begin{aligned}
C(\mathbf{r}) &= \int_{0}^{\infty} c(\mathbf{r}(t), \mathbf{d}) \sigma(\mathbf{r}(t)) T(t) \, dt,\\
T(t) &= e^{- \int_{0}^{t} \sigma(\mathbf{r}(s)) \, ds}
\end{aligned}
\label{eq:volrender}
\end{equation}
This equation is the basis for recent Novel View Synthesis approaches like NeRF and 3D Gaussian Splatting.

\textbf{NeRF and Scene Representation:} NeRF optimizes a neural network representing the fields \(c\) and \(\sigma\) using training images and their corresponding camera poses \cite{Neural_Radiance_Fields}. During training, images are rendered by ray-casting from training poses and compared to ground truth using a loss function, allowing the update of network parameters via gradient descent. During inference, the trained network generates realistic images from new camera poses by ray casting.

More precisely, NeRF uses the differentiable Volume Ray Casting algorithm inspired by Max computations in \cite{MaxOpt}. It consists in launching rays in the 3D space to compute a discretized version of equation \ref{eq:volrender} using  \(N\) samples \(\tilde{t_i} \) along the ray:

\begin{equation}
\begin{aligned}
C(\mathbf{r}) &= \sum_{i=0}^{N} \left(1 - \exp(-\sigma_i \Delta t)\right) c_i T_i \\
T_i &= \exp \left( -\sum_{j=0}^{i-1} \sigma_j \Delta t \right)
\label{eq:vol_raycasting}
\end{aligned}
\end{equation}
where \(\Delta t\) the discretization step, and the \(i\)-th field values denoted as \(\sigma_i = \sigma(\mathbf{r}(\tilde{t_i}))\) and \(c_i = c(\mathbf{r}(\tilde{t_i}),\mathbf{d})\), this amounts to assuming that the fields are piecewise constant in samples neighborhood. The reader interested in the computational details can find more information in 
\cite{MaxOpt}\cite{Max2010LocalAG}. 

NeRF's key contribution is using a differentiable volumetric rendering algorithm to learn scene parameters with a neural network. Later research improved scene representation by using different structures to locally store parameters, reducing reliance on a costly global neural network and enhancing appearance through better local parameter adaptation. Moreover, due to their sparse nature, some data structures can apply classic ray-casting acceleration strategies such as empty space skipping \cite{Hadwiger2018SparseLeap}.  Among these structures, one can notably mention voxel grids \cite{Direct_Voxel_Grid_optimization}, grids with hash encoding \cite{Instant_NGP}, grids of small MLPs \cite{KiloNeRF}, sets of tetrahedrons \cite{Tetra_NeRF}, and point clouds \cite{Point_NeRF}.  Our hypothesis is that point clouds are best suited to adapt to scenes. Each point can host a locally supported function, such as Gaussian functions \cite{3D_Gaussian_Splatting}, representing local scene details. Optimizing these functions' positions and shapes enables precise adaptation to scene geometry, unlike voxel grids constrained by their resolution. Furthermore, to the best of our knowledge, no method currently allows for optimizing a scene represented only by irregularly spaced basis functions with the ray-casting algorithm without coarse approximations. The approach in Point-NeRF \cite{Point_NeRF} is interesting but requires using a MVSNet, which proves cumbersome. Additionally, its expression of \(c\) and \(\sigma\) involves using a neural network and only considers the eight closest neighbors in its calculations. The NeuRBF approach \cite{chen2023neurbf} is also promising, as it uses a hybrid representation with adaptive RBFs for fine details and grid-based RBFs for efficient continuous representation. The use of grid-based RBFs is needed because the algorithm relies on a limited number of nearest adaptive RBFs (5 for Blender dataset), causing discontinuities. Like Point-NeRF, this limits its ability to fully exploit the approximation power of sparse basis functions. Moreover, as discussed in the following section, 3D Gaussian Splatting and numerous point cloud rendering methods \cite{Differentiable_Point_Based} make the splatting approximation, which can cause visible artifacts such as flickering when changing viewpoints. This choice is due to the scarcity of algorithms that efficiently perform ray casting on local basis functions associated with points.

\textbf{Gaussian Splatting:} 3D Gaussian Splatting \cite{3D_Gaussian_Splatting} made a breakthrough by using a differentiable splatting algorithm to render a scene represented by Gaussians. It draws inspiration from the classic Elliptical Weighted Average (EWA) Splatting approach \cite{ewa}. The main advantages of this approach are the use of Gaussians, which allow for efficient adaptation to the scene geometry, and an efficient tile-based rasterizer, which enables fast rendering of the scene. Starting from a set of 3D Gaussians associated with color parameters $c_i$ and opacity $\alpha_i$, the splatting algorithm projects the 3D Gaussians into the camera space, performs a global sorting by their depth, and sequentially adds their contribution from the nearest to the farthest for each pixel color \(C\) using the equation:
\begin{equation}
\begin{aligned}
C &= \sum_{i \in \mathcal{N}} c_i \alpha'_i \prod_{j=1}^{i-1} (1 - \alpha'_j) \\
\alpha'_i &= \alpha_i \times \exp \left( -\frac{1}{2} (x' - \mu'_i)^\top \Sigma'^{-1}_i (x' - \mu'_i) \right)
\end{aligned}
\label{eq:splatting}
\end{equation}
Where \(x'\) the pixel coordinates in the camera frame, \(\mu'_i\) and \(\Sigma'_i\) are the mean and covariance of the Gaussian projected into the camera frame, and \(\mathcal{N}\) denotes the set of Gaussians. More implementation details are available in the original paper \cite{3D_Gaussian_Splatting} and the following review \cite{chen2024survey3dgaussiansplatting}. Like the discretized equation of ray casting, the splatting formulation \ref{eq:splatting} is also an approximation of  Eq. \ref{eq:volrender}. However, the approximations used differ: Gaussians are assumed to be non-overlapping, which does not hold for complex surfaces. Indeed, multiple overlapping Gaussians are required for precise scene geometry reconstruction. Then, each Gaussian contributes at most once per ray, aggregated at a distance corresponding to the depth of the Gaussian (Gaussian mean projected along the viewing direction). Also, the sequential addition of Gaussian contributions based on global depth is an approximation. It does not account for the proper intersection between the Gaussians and a given ray, further reducing the coherence of the representation. Finally, the projection of Gaussians into the camera frame is also an approximation introduced in \cite{ewa}. Thus, we use the ray casting algorithm to suffer fewer approximations because we hypothesize that, given a high enough sampling rate, it maintains better coherence in scene representation during training and inference and should yield better graphical results.

\begin{figure*}[ht]
  \centering
   \includegraphics[trim={0cm 1cm 1cm 0cm},clip,width=0.8\linewidth]{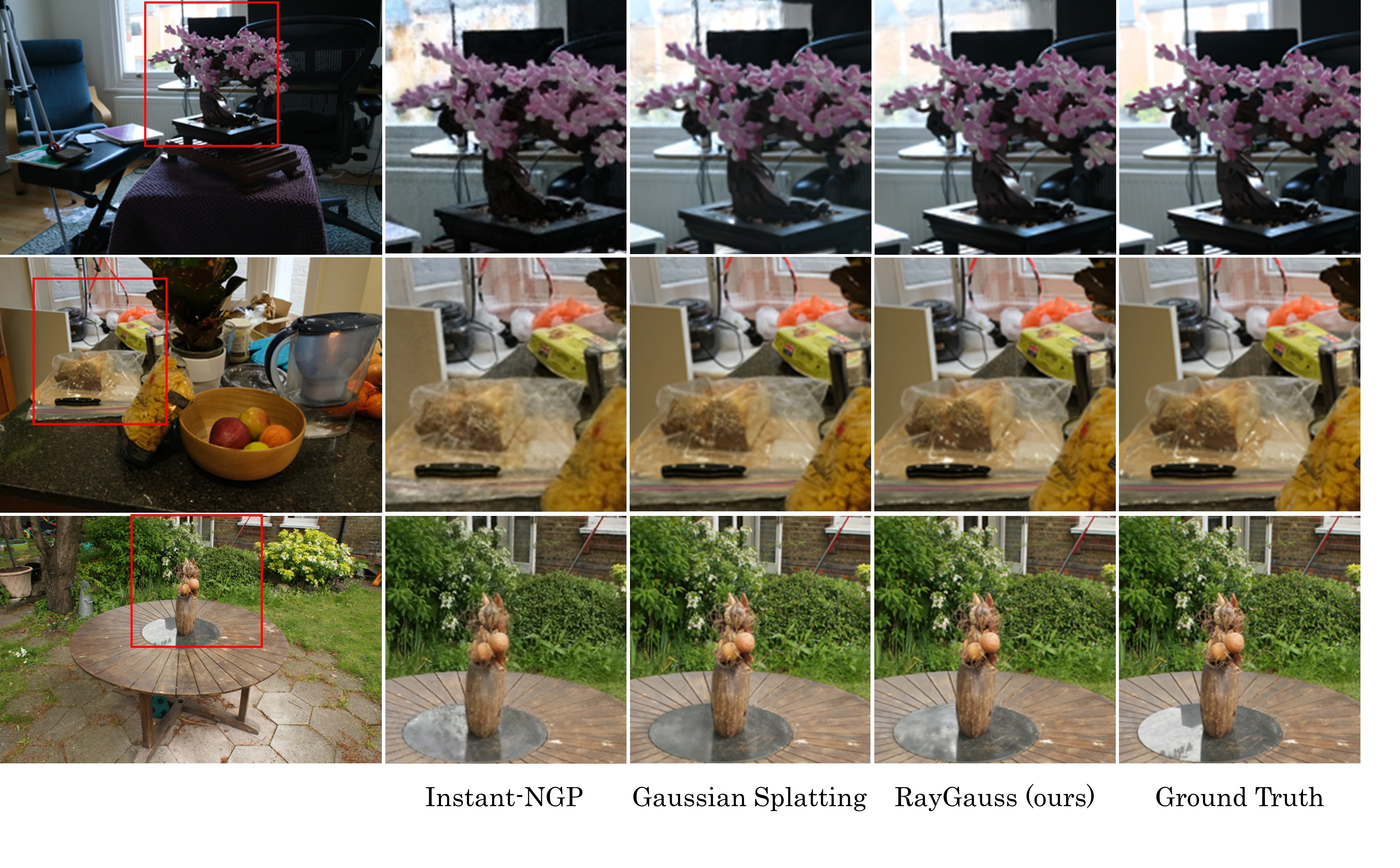}
   \caption{\textbf{Test set images from the Mip-NeRF 360 Dataset}}
   \label{fig:test_mipnerf360}
\end{figure*}

\section{Scene Representation}

Our rendering model is based on Eq. \ref{eq:vol_raycasting}. Thus, the performance of our approach greatly depends on the formulation of our functions \( \sigma \) and \( c \). The main idea of our work is to approximate these fields through a decomposition using irregularly spatially distributed basis functions while maintaining physical coherence. This will allow for better adaptation to the geometry of the represented scene without resolution limits. Thus, for a set of basis functions \( \{\phi_l,\psi_l\}_{l=1,..,N} \), our fields will be of the form:
\begin{equation}
\begin{aligned}
\sigma(\mathbf{x}) &= \sum_{l=1}^{N} \tilde{\sigma_l} \phi_l(\mathbf{x}) \\
c(\mathbf{x},\mathbf{d}) &= \sum_{l=1}^{N} c_l({\mathbf{d}}) \psi_l(\mathbf{x})
\end{aligned}
\label{eq3}
\end{equation}
In what follows, we will explain the choices of weights and basis functions we made to best approximate the scene with the fields \( \sigma \) and \( c \) while maintaining physical consistency in the \( \sigma \)/\( c \) relationship.

\subsection{Irregularly Spatially-Distributed Emissive and Absorptive Primitives}
As we will see, the expression previously given in Equation \ref{eq3} can be physically assimilated to the case of a scene composed of \( N \) independent emissive and absorbing entities, each distributed over the support of their basis function with their dedicated density \( \sigma_{l}\) and emissive radiance \( c_{l}\)\cite{Stamnes_Thomas_Stamnes_2017}\cite{RadTransfinStelAtmo}. Indeed, given a basis function \(\phi_l\) for \(l \in [1, N]\), the evolution of the density \( \sigma_l\) of the \(l\)-th entity in the scene is such that:
\begin{equation}
\sigma_l(\mathbf{x}) = \tilde{\sigma_l} \cdot \phi_l(\mathbf{x})
\end{equation}
where \(\tilde{\sigma_l}\) is the maximum density with \(0 < \phi_l(\mathbf{x}) < 1\). 
On the other hand, we choose an emitted radiance \( c \) that depends only on the direction of emission \( \mathbf{d} \) and is constant in space. 
\begin{equation}
c_l(\mathbf{x},\mathbf{d}) = c_l({\mathbf{d}})
\end{equation}
This representation allows us to depict objects whose appearance changes according to the viewing direction. We will develop this aspect further.

Then, to derive a physically coherent total density \( \sigma \) and total emitted radiance term \( c \) for the scene, we can revisit the radiative transfer equation \ref{eq:rad_transfer} for the case of \( N \) independent entities with distinct density \( \sigma_{l}(\mathbf{x}) \) and emissive radiance \( c_{l} \), in which case it is expressed as follows:

\begin{equation}
(\mathbf{d} \cdot \nabla)L(\mathbf{x}, \mathbf{d}) = -\left(\sum_{l=1}^{N} \sigma_{l}(\mathbf{x})\right) L(\mathbf{x}, \mathbf{d}) + \sum_{l=1}^{N} \sigma_{l}(\mathbf{x}) c_{l}
\end{equation}

where: \( (\mathbf{d} \cdot \nabla)L(\mathbf{x}, \mathbf{d}) \) is the rate of change of radiance \( L(x, \mathbf{d}) \)\cite{Stamnes_Thomas_Stamnes_2017} \cite{RadTransfinStelAtmo}. The previous radiative transfer equation allows for considering variations in radiance due to absorption phenomena (first term) and emission phenomena (second term) due to each independent entity. We recall that the radiative transfer equation for a medium described by global density function \(\sigma(\mathbf{x})\) and global emitted radiance field \(c(\mathbf{x},\mathbf{d})\) is given by Eq. \ref{eq:rad_transfer}.
Thus, we deduce that it is possible to place ourselves within this framework by defining the density field as follows:
\begin{equation}
\sigma(\mathbf{x}) = \sum_{l=1}^{N} \sigma_l(\mathbf{x}) = \sum_{l=1}^{N} \tilde{\sigma_l} \phi_l(\mathbf{x})
\end{equation}
Furthermore, the emitted radiance field can then be defined as:
\begin{equation}
c(\mathbf{x}, \mathbf{d})  = \dfrac{ \sum_{l=1}^{N} c_l({\mathbf{d}}) \tilde{\sigma_l} \phi_l(\mathbf{x})}{\sum_{m=1}^{N} \tilde{\sigma}_m \phi_m(\mathbf{x})}
\end{equation}
As initially intended, we find a formulation with a weighted sum of basis functions. Moreover, this formulation allows us to maintain consistent physical behavior during rendering, avoiding some visual artifacts while using a flexible basis function.

\subsection{Selection of the basis function}
The base function class must allow our primitives to adapt optimally to the scene's geometry. In particular, we study radial and elliptical basis functions to leverage their approximation power\cite{rbfpower}. Thus, each basis function $\phi_l$ is centered at $\mu_l$ and evolves with $r = d(\mathbf{x}, \mu_l)$, where $d$ is the Euclidean distance for radial functions or the Mahalanobis distance for elliptical functions. Moreover, we restrict ourselves to decreasing functions with respect to \(r\), allowing us to use point clouds representing a raw depiction of the scene as an initialization for the center \(\mu_l\) of the basis functions. Indeed, the positions of the points provide a nice prior for regions that highly interact with light.

More specifically, we have chosen anisotropic Gaussian functions as our basis function class, as this gives the best results among the functions studied (see Tab. \ref{tab:ablation}).  A given basis function can thus be expressed as follows:
\begin{equation}
    \mathcal{G}(\mathbf{x}; \mathbf{\mu}, \mathbf{\Sigma}) = \exp \left( -\frac{1}{2} (\mathbf{x} - \mathbf{\mu})^T \mathbf{\Sigma}^{-1} (\mathbf{x} - \mathbf{\mu}) \right)
\end{equation}
Where \( \mathbf{\mu} \) is the mean vector, \( \mathbf{\Sigma} \) is the covariance matrix. Here, we can notice that \(\sqrt{(\mathbf{x} - \mathbf{\mu})^T \mathbf{\Sigma}^{-1} (\mathbf{x} - \mathbf{\mu}) }\) is the Mahalanobis distance associated with the matrix \(\Sigma\). To achieve a more suitable parameterization, the covariance matrix \(\mathbf{\Sigma}\), being positive and symmetric, will be decomposed according to the spectral theorem into a rotation matrix \(\mathbf{R}\) and a scaling matrix \(\mathbf{S}\) such that $\mathbf{\Sigma} = \mathbf{R} \mathbf{S} \mathbf{S}^T  \mathbf{R}^T$. The rotation \(\mathbf{R}\), as part of the covariance matrix $\mathbf{\Sigma}$ decomposition, can be expressed in terms of unit quaternions \( \mathbf{q} \in \mathbb{R}^4 \) such that \( \mathbf{R} = \mathbf{R}(\mathbf{q}) \), and the scaling matrix \( \mathbf{S} \) can be expressed in terms of \( \mathbf{s}=(s_x, s_y, s_z) \) such that \( \mathbf{S} = \text{diag}(s_x, s_y, s_z) \). Thus, we will refer to our Gaussian basis functions as \( \mathcal{G}(\mathbf{x}; \mathbf{\mu}, \mathbf{q}, \mathbf{s}) \). We can now describe our density field \(\sigma(\mathbf{x})\) and emitted radiance field \( c(\mathbf{x}, \mathbf{d})  \) as follows:
\begin{equation}
\begin{aligned}
    \sigma(\mathbf{x}) &= \sum_{l=1}^N \tilde{\sigma}_l \mathcal{G}(\mathbf{x}; \mu_l, \mathbf{q}_l, \mathbf{s}_l) \\
    c(\mathbf{x}, \mathbf{d}) &= \frac{\sum_{l=1}^N c_l({\mathbf{d}}) \tilde{\sigma}_l \mathcal{G}(\mathbf{x}; \mu_l, \mathbf{q}_l, \mathbf{s}_l)}{\sum_{m=1}^N \tilde{\sigma}_m \mathcal{G}(\mathbf{x}; \mu_m, \mathbf{q}_m, \mathbf{s}_m)}
\end{aligned}
\end{equation}

\subsection{Direction-dependent emitted radiance}
To correctly represent objects whose color appearance varies by observation direction, we made the emitted radiance of our primitives dependent on the viewing direction. One parameterization that gave the best results was a combination of spherical harmonics and spherical Gaussians, such that emitted radiance of l-th primitive is:
\begin{equation}
    c_l(\mathbf{d})=c_{\text{low},l}(\mathbf{d}) + c_{\text{high},l}(\mathbf{d})
\end{equation}
where:
\begin{equation}
\begin{aligned}
    c_{\text{low},l}(\mathbf{d}) &= \sum_{j=0}^{L_1} \sum_{m=-j}^{j} \tilde{c}_{l,jm} Y_{jm}(\mathbf{d})\\
    c_{\text{high},l}(\mathbf{d}) &=\sum_{j=0}^{L_2} k_{l,j} e^{\lambda_{l,j}(\mathbf{d} \cdot \mathbf{p}_{l,j} -1)}
\end{aligned}
\end{equation}
\(Y_{jm}(\mathbf{d})\) is the spherical harmonic of degree \(j\) and order \(m\) associated with its coefficient \(\tilde{c}_{l,jm}\in \mathbb{R}^3\), while \(k_{l,j}\in \mathbb{R}^3\) is the coefficient of the \(j\)-th spherical gaussian with lobe sharpness \(\lambda_{l,j}\in \mathbb{R}\) and lobe direction \(\mathbf{p}_{l,j}\in \mathbb{R}^3\).

This choice is motivated by low-degree Spherical Harmonics being better suited to represent low-frequency variations. In contrast, Spherical Gaussians can represent any frequency, including high frequencies. Thus, their combined use allows for representing both low and high-frequency phenomena. Similar considerations can be found for different applications in articles such as MixLight~\cite{ji2024mixlightborrowingbestspherical}.

\section{Scene initialization and optimization}
\begin{figure*}[ht]
  \centering
   \includegraphics[width=\linewidth]{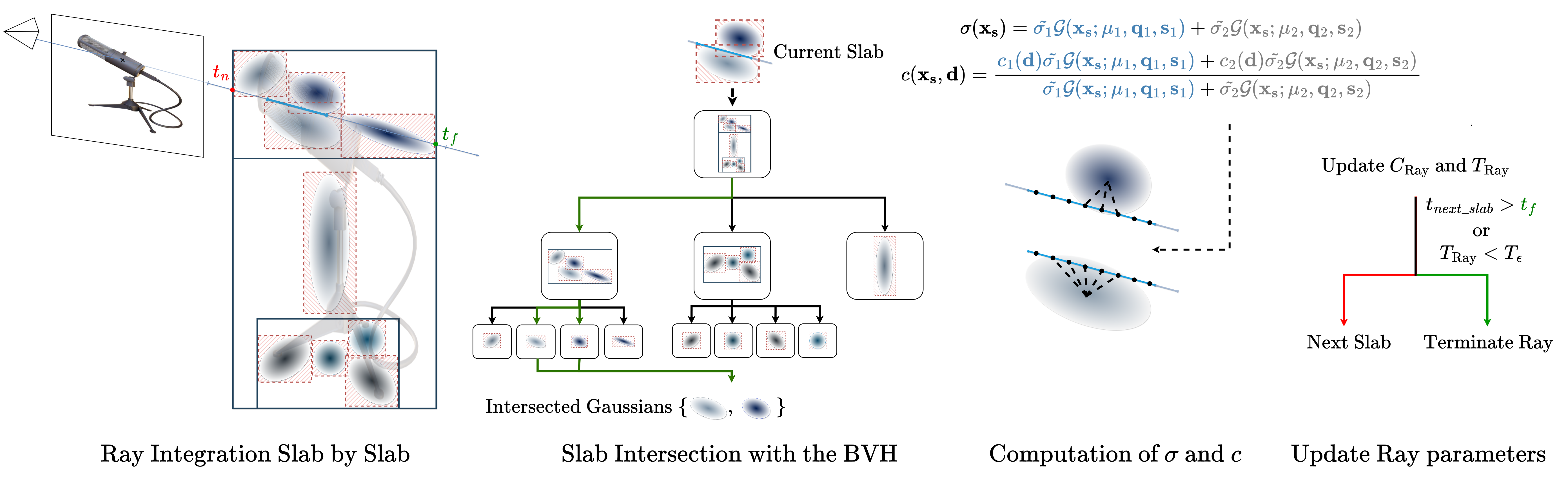}
   \caption{\textbf{Overview of our slab-by-slab Ray Casting algorithm:} When a given ray intersects the scene's bounding box, the ray is processed by successive slabs. For a given slab, we check the intersection with the Gaussians using the BVH. Then, the attributes of the slab are calculated to update the current color and transmittance of the ray. }
   \label{fig:slab_ray_casting}
\end{figure*}
Our goal is to use the scene representation discussed earlier and optimize its parameters \( P = \{ (\tilde{\sigma}_l, \tilde{c}_l, \mu_l, \mathbf{q}_l, \mathbf{s}_l) \mid l = 1, \ldots, N \} \), where \( \tilde{c}_l \) denotes the set of colorimetric parameters of the l-th basis function, \(\tilde{\sigma}_l\) its density parameter, \(\mu_l, \mathbf{q}_l, \mathbf{s}_l\) the parameters determining the shape of the base function. Once the parameters are optimized, the objective is to infer images that faithfully represent the scene.
\subsection{Scene initialization}
To initialize this representation, we leverage sparse point clouds obtained via Structure from Motion (SfM), where the position and color of these points are used to initialize colorimetric parameters and the center of the basis functions \( \mu_l \). Additionally, unit quaternions are initialized to match the identity matrix, and the scale parameters are initialized to the average distance from each point to its three nearest neighbors.
\subsection{Scene Optimization}
After initialization, we optimize the representation using stochastic gradient descent, relying on training images paired with their camera poses. Iteratively, over the training set, we generate the scene image from the current camera pose and compare it to the current training image using a given loss function. We then compute loss gradients with respect to scene parameters for backpropagation updates. In particular, the loss function \( L \) is a weighted sum of the \( L_{1} \) loss function and the structural dissimilarity (DSSIM):

\[
L =  (1 - \lambda)\cdot L_{1} + \lambda \cdot \text{DSSIM}
\]
where \( \lambda \) is a weighting factor between the L1 loss and DSSIM, we follow best practices developed in 3D Gaussian Splatting \cite{3D_Gaussian_Splatting}. Furthermore, we are in a similar case to 3D Gaussian Splatting regarding adaptive control of primitives. Indeed, we also deal with primitives irregularly distributed in space. For this reason, we rely on the point cloud densification heuristics described in \cite{3D_Gaussian_Splatting}. Additionally, we remove outlier points with density below a fixed threshold \( \sigma_{\epsilon} \) and progressively unlock colorimetric parameters, starting with spherical harmonics from low to high degree and then spherical Gaussians, to enforce model coherence from low to high frequencies.

\section{Spatially irregularly distributed base functions ray casting}
This approach has been scarcely explored due to the major obstacle of implementing an algorithm that enables efficient differentiable volume ray casting for a scene defined with irregularly distributed basis functions. Firstly, since the support of our basis functions is unbounded in the chosen case (Gaussian functions), it would potentially require evaluating all basis functions to compute \( \sigma \) and \( c \) at any spatial position. This promises to slow down the rendering of our images significantly. To mitigate this, we truncate our basis functions to a local domain. The criterion retained to determine this domain for each primitive is as follows:

\begin{equation}
    \sigma_l(\mathbf{x})_{\text{approx}}=
\begin{cases}
    \sigma_l(\mathbf{x}) & \text{if } \sigma_l(\mathbf{x}) \geq \sigma_{\epsilon} \\
    0 & \text{if } \sigma_l(\mathbf{x}) < \sigma_{\epsilon}
\end{cases}
\end{equation}
where \( \sigma_{\epsilon} \) is a fixed threshold, this amounts to disregarding areas with a density coefficient below a given threshold. These areas have minimal interaction with light and, therefore, have little influence on the resulting color. Thus, for a low threshold, this approximation appears acceptable.
Now that the influence of each primitive is confined to a specific domain, we implement a differentiable algorithm capable of computing an image through volume ray casting with radial or elliptical basis functions truncated to a finite domain. We rely on a bounding volume hierarchy (BVH) to achieve this, drawing inspiration from the work conducted by \cite{Knoll2014RBFVR}. Thus, to compute the color associated with a given ray, we construct a BVH containing ellipsoids if the chosen base is elliptical or spheres if it is radial. These geometric figures represent the level set associated with the previous truncation. Once the acceleration structure has been built, we launch each ray in parallel in the acceleration structure, in practice using the Nvidia OptiX framework. Integration along the ray is done sequentially by slabs of samples, each containing a fixed number of samples (see Fig. \ref{fig:slab_ray_casting}). 
By doing this, our approach avoids the drawbacks of naive methods: integrating sample-by-sample is slow due to repeated neighborhood calculations. While, calculating $\sigma$ and $c$ attributes for all samples in one pass limits the number of samples, requires a large buffer, and prevents early ray termination. Our method sequentially calculates neighborhoods for entire slabs and stores slab attributes in a small fixed-size buffer, reducing costly calculations and allowing efficient early ray termination at the slab level.


\section{Implementation and Experiments}
\label{sec:experiments}
\begin{figure*}[ht]
  \centering
   \includegraphics[trim={0cm 0.7cm 0cm 0cm},clip,width=0.8\linewidth]{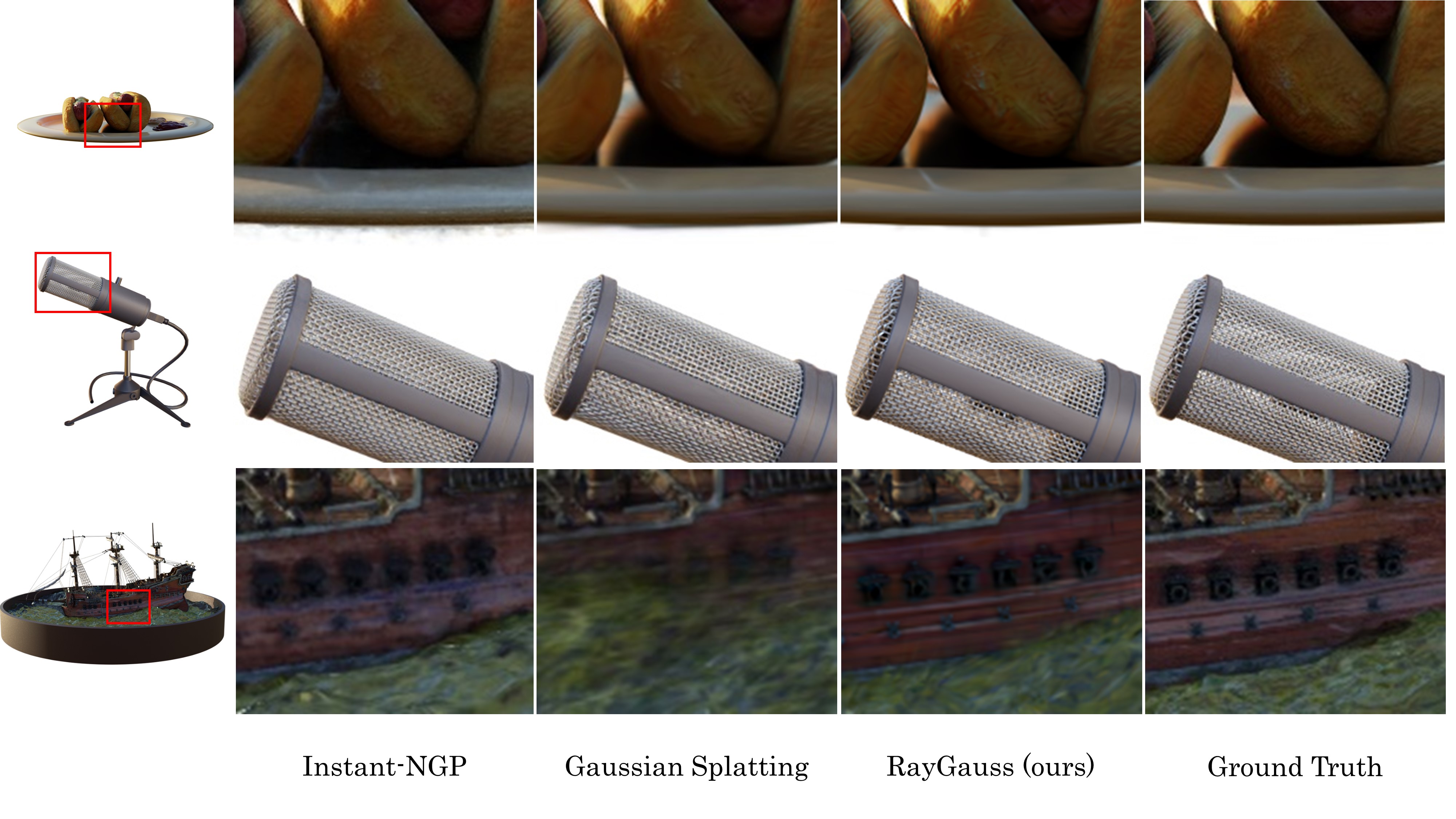}
   \caption{\textbf{Test set images from the Blender Dataset}}
   \label{fig:test_blender_images}
\end{figure*}

\subsection{Implementation}
Our implementation relies on the PyTorch library for managing the optimization framework, combined with a Python binding of Nvidia Optix \cite{Parker10OptiX} using the Cupy library. Thus, the computationally intensive parts of our algorithm, including BVH management, ray casting, and derivative calculations, are implemented with the Optix library and custom CUDA kernels. Details on our optimization parameters can be found in supplementary materials.

\subsection{Results and Evaluation}
To quantify our method's performance, we tested it on two widely used NVS datasets: the Blender dataset \cite{Neural_Radiance_Fields} and the Mip-NeRF 360\degree  dataset \cite{Mip_NeRF_360}. This allows us to test our method across a wide range of situations. The Blender dataset includes 8 scenes with complex objects exhibiting non-Lambertian effects, each with 100 training images, 200 test images, and exact camera parameters for 360$\degree$ views.
The Mip NeRF 360$\degree$ dataset comprises 9 scenes: 5 outdoor and 4 indoor, each featuring a central object with a complex background. Each scene includes images with camera parameters estimated by COLMAP \cite{schoenberger2016sfm}. We evaluate our method by downscaling images by a factor of 8 for this dataset due to large image size. Moreover, we compare our results mainly with 3D Gaussian Splatting \cite{3D_Gaussian_Splatting} and subsequent research papers such as Mip-Splatting \cite{Yu2023MipSplatting} due to the similarity in the representation. The results in Tables \ref{tab:blender} and \ref{tab:mipnerf360} are from state-of-the-art methods, except those marked with *, which were recalculated using available code.

\textbf{Results on the Blender dataset:}
For synthetic scenes in the Blender dataset, we provide quantitative comparisons in Tab. \ref{tab:blender} and qualitative comparisons in Fig. \ref{fig:test_blender_images} with state-of-the-art methods. Quantitative results show that our approach surpasses state-of-the-art results on most individual scenes. Additionally, our average PSNR of \textbf{34.53 dB} highlights our model's superior adaptation to scene geometry and appearance compared to other state-of-the-art methods. Qualitative results also support this, for example, with a boat hull better reconstructed in the \textit{ship} scene compared to 3D Gaussian Splatting reconstruction (Figure~\ref{fig:test_blender_images}). NeuRBF \cite{chen2023neurbf} performs well, similar to our method, but has an inference time of a few seconds per image—about 100x longer than ours (25 FPS on Blender dataset, see Tab. \ref{tab:performance})—and does not support the Mip-NeRF360 dataset.

\begin{table}[h]
  \centering
  \resizebox{\columnwidth}{!}{%
  \begin{tabular}{l | c c c c c c c c | c }
    & Chair & Drums & Ficus & Hotdog & Lego & Materials & Mic & Ship & \textbf{Avg.}\\
    \hline
NeRF~\cite{Neural_Radiance_Fields} & 34.17 & 25.08 & 30.39 & 36.82 & 33.31 & 30.03 & 34.78 & 29.30 & 31.74 \\
Zip-NeRF~\cite{zipnerf} & 34.84 & 25.84 & 33.90 & 37.14 & 34.84 & \cellcolor{orange!40}31.66 & 35.15 & \cellcolor{yellow!40}31.38 & 33.10 \\
Instant-NGP~\cite{Instant_NGP} & 35.00 & 26.02 & 33.51 & 37.40 & 36.39 & 29.78 & 36.22 & 31.10 & 33.18 \\ 
Mip-NeRF360~\cite{Mip_NeRF_360} & 35.65 & 25.60 & 33.19 & 37.71 & 36.10 & 29.90 & \cellcolor{yellow!40}36.52 & 31.26 & 33.24 \\
Point-NeRF~\cite{Point_NeRF} & 35.40 & 26.06 & \cellcolor{red!40}36.13 & 37.30 & 35.04 & 29.61 & 35.95 & 30.97 & 33.30 \\
Gaussian Splatting~\cite{3D_Gaussian_Splatting}* & 35.85 & 26.22 & 35.00 & 37.81 & 35.87 & 30.00 & 35.40 & 30.95 & 33.39 \\
Mip-Splatting~\cite{Yu2023MipSplatting}* & 36.03 & \cellcolor{yellow!40}26.29 & \cellcolor{orange!40}35.33 & \cellcolor{yellow!40}37.98 & 36.03 & 30.29 & 35.63 & 30.50 & 33.51 \\
PointNeRF++~\cite{sun2023pointnerfpp} & \cellcolor{yellow!40}36.32 & 26.11 & 34.43 & 37.45 & \cellcolor{yellow!40}36.75 & 30.32 & \cellcolor{orange!40}36.85 & 31.34 & \cellcolor{yellow!40}33.70 \\
NeuRBF~\cite{chen2023neurbf}*& \cellcolor{orange!40}36.54 & \cellcolor{orange!40}26.38 & 35.01 & \cellcolor{red!40}38.44 & \cellcolor{red!40}37.35 & \cellcolor{red!40}34.12 & 36.16 & \cellcolor{orange!40}31.73 & \cellcolor{orange!40}34.47 \\
    \hline
RayGauss (ours) & \cellcolor{red!40}37.20 & \cellcolor{red!40}27.14 & \cellcolor{yellow!40}35.11 & \cellcolor{orange!40}38.30 & \cellcolor{orange!40}37.10 & \cellcolor{yellow!40}31.36 & \cellcolor{red!40}38.11 & \cellcolor{red!40}31.95 & \cellcolor{red!40}34.53\\
  \end{tabular}
  }
  \caption{\textbf{PSNR scores for Blender dataset.}}
  \label{tab:blender}
\end{table}

\textbf{Results on the Mip-NeRF 360 Dataset:} Regarding Mip-NeRF 360, we provide quantitative results in Tab. \ref{tab:mipnerf360} and qualitative comparisons in Fig. \ref{fig:test_mipnerf360}. A general trend is that our approach performs better on most indoor scenes than state-of-the-art methods like Mip-Splatting \cite{Mip_NeRF_360}, and slightly worse on outdoor scenes. This may be due to the adaptive control of gaussians strategy, which adapts well in Gaussian Splatting \cite{3D_Gaussian_Splatting} but may perform worse with our approach. Nevertheless, we achieved better overall results than Mip-Splatting and 3D Gaussian Splatting, with an average PSNR of 29.71 dB, \textbf{+0.56 dB} higher than 3D Gaussian Splatting. Zip-NeRF performs slightly better on Mip-NeRF 360 but requires over 7 hours of training per scene and about 1.5 seconds per image for inference, which is significantly higher than our approach (see Tab. \ref{tab:performance}).

\begin{table}[h!]
  \centering
  \resizebox{\columnwidth}{!}{%
  \begin{tabular}{l | c c c c | c c c c c | c }
  & \multicolumn{4}{|c|}{\textbf{Indoor}} & \multicolumn{5}{|c|}{\textbf{Outdoor}} \\
    & bonsai & counter & kitchen & room & bicycle & flowers & garden & stump & treehill & \textbf{Avg.}\\
    \hline
NeRF~\cite{Neural_Radiance_Fields}* & 22.10 & 22.34 & 22.00 & 24.46 & 19.35 & 19.49 & 22.70 & 21.43 & 21.02 & 21.65  \\
Instant-NGP~\cite{Instant_NGP}* & 27.04 & 24.25 & 23.44 & 27.30 & 23.69 & 21.41 & 25.64 & 22.56 & 22.22 & 24.17 \\
Gaussian Splatting~\cite{3D_Gaussian_Splatting}* & 33.42 & \cellcolor{yellow!40}30.21 & \cellcolor{yellow!40}33.40 & \cellcolor{yellow!40}32.95 & \cellcolor{yellow!40}27.33 & \cellcolor{yellow!40}23.71 & 29.58 & \cellcolor{orange!40}27.78 & 24.00 & 29.15\\
Mip-Splatting~\cite{Yu2023MipSplatting}* & \cellcolor{yellow!40}33.44 & \cellcolor{orange!40}30.43 & \cellcolor{orange!40}34.30 & \cellcolor{orange!40}33.30 & \cellcolor{orange!40}27.62 & \cellcolor{orange!40}23.79 & \cellcolor{yellow!40}29.78 & \cellcolor{red!40}27.89 & \cellcolor{yellow!40}24.25 & \cellcolor{yellow!40}29.42 \\
Zip-NeRF~\cite{zipnerf}* & \cellcolor{red!40}36.10 &    30.13 &    32.85 &    \cellcolor{red!40}34.20 &    \cellcolor{red!40}28.10 &    \cellcolor{red!40}24.25 &    \cellcolor{red!40}30.24 &    \cellcolor{orange!40}27.78 &    \cellcolor{red!40}25.72 & \cellcolor{red!40}29.93 \\
\hline
RayGauss (ours) & \cellcolor{orange!40}35.22 & \cellcolor{red!40}31.79 & \cellcolor{red!40}35.43 & \cellcolor{yellow!40}32.95 & 27.21 & 23.53 & \cellcolor{orange!40}29.91 & \cellcolor{yellow!40}27.13 & \cellcolor{orange!40}24.26 & \cellcolor{orange!40}29.71 \\
  \end{tabular}
  }
  \caption{\textbf{PSNR scores for Mip-NeRF 360 dataset.} All methods are trained and tested on downsampled images by a factor of 8.}
  \label{tab:mipnerf360}
\end{table}

\textbf{Performance Analysis:} We discuss here results in Tab. \ref{tab:performance}, which show the performance of our approach compared to Gaussian Splatting. These metrics were obtained using an RTX 4090 GPU with 24GB of RAM on 800x800 pixel images. Our approach has slower training and inference times compared to Gaussian Splatting due to the computational cost of ray casting. Generally, our computation times depend mainly on the number of rays, while Gaussian Splatting's times are influenced by the number of projected primitives.
However, our training times remain reasonable (30 minutes per scene on Blender) and real-time inference is achieved on Blender, although slower on Mip-NeRF 360. We also note that our method generates fewer primitives for better visual quality. Thus, Gaussian Splatting likely adds more Gaussians to compensate for splatting artifacts.


\begin{table}[h]
  \centering
  \resizebox{\columnwidth}{!}{%
  \begin{tabular}{l|c|c|c}
    \textbf{} & Number of Gaussians & Training Time & Inference Time \\
    \hline
    \multicolumn{4}{c}{\cellcolor[HTML]{FFFFFF}\textbf{Blender dataset}} \\
    \hline
    Gaussian Splatting~\cite{3D_Gaussian_Splatting}* & 295k & 10 min & 400 FPS \\
    RayGauss (ours) & 205k & 30 min & 25 FPS \\
    \hline
    \multicolumn{4}{c}{\cellcolor[HTML]{FFFFFF}\textbf{Mip-NeRF 360}} \\
    \hline
    Gaussian Splatting~\cite{3D_Gaussian_Splatting}* & 2.7M & 15 min & 130 FPS \\
    RayGauss (ours) & 2.3M & 1h 30 min& 5 FPS \\
  \end{tabular}%
  }
  \caption{\textbf{Comparison of Gaussian Splatting and RayGauss performance on Blender and Mip-NeRF 360 datasets.} Numbers are averaged per scene on each dataset.}
  \label{tab:performance}
\end{table}

\subsection{Ablation Studies}
\begin{table}[h]
  \centering
  \resizebox{\columnwidth}{!}{%
  \begin{tabular}{l || c | c }
    Method \textbackslash Metrics& PSNR ↑ & SSIM ↑\\
    \hline
    \hline
    (Gaussian Splatting~\cite{3D_Gaussian_Splatting}) Spherical Harmonics* & \cellcolor{yellow!40}33.39 & \cellcolor{yellow!40}0.969 \\
    (Ours) Spherical Harmonics & \cellcolor{red!40}33.75 & \cellcolor{red!40}0.972 \\
    \hline
    \hline
    (Ours) Isotropic Gaussian & \cellcolor{yellow!40}33.44 & \cellcolor{yellow!40}0.969 \\
    (Ours) Anisotropic Gaussian & \cellcolor{red!40}34.53 & \cellcolor{red!40}0.974 \\
    \hline
    \hline
    (Ours) Anisotropic Inverse Multiquadric & 30.84 & 0.962\\
    (Ours) Anisotropic Inverse Quadratic & 32.95 & \cellcolor{yellow!40}0.967\\
    (Ours) Anisotropic $C^0$ Matérn & 33.83& \cellcolor{orange!40}0.972 \\
    (Ours) Anisotropic Bump & \cellcolor{orange!40}34.18& \cellcolor{orange!40}0.972\\
    (Ours) Anisotropic Wendland &  \cellcolor{yellow!40}34.09 & \cellcolor{orange!40}0.972 \\
    (Ours) Anisotropic Gaussian & \cellcolor{red!40}34.53 & \cellcolor{red!40}0.974\\
    \hline
    \hline
    (Ours) RGB & 30.39 & 0.961  \\
    (Ours) Spherical Harmonics & \cellcolor{yellow!40}33.75 & \cellcolor{yellow!40}0.972 \\
    (Ours) Spherical Gaussians & \cellcolor{orange!40}34.40 & \cellcolor{orange!40}0.973 \\
    (Ours) Spherical Gaussians + Spherical Harmonics & \cellcolor{red!40}34.53 & \cellcolor{red!40}0.974 \\
  \end{tabular}
  }
  \caption{\textbf{Evaluation of different aspects of our method:} Rendering Algorithm, Basis Function, and Colorimetric Parameters by averaging PSNR and SSIM scores on Blender Dataset}
  \label{tab:ablation}
\end{table}
In this section, we quantify the influence of the different components of our approach. Thus, we test different configurations by averaging our PSNR and SSIM results on the NeRF-Synthetic dataset. Results are summarized in Tab. \ref{tab:ablation}. Additionally, we prioritize PSNR analysis to assess the different approaches, as this metric has been found more effective than SSIM or LPIPS according to recent studies \cite{Perceptual_Quality_Assessment}.

\textbf{3D Gaussian Splatting:} We first compare our method with Gaussian basis function and colorimetric parameters similar to 3D Gaussian Splatting (Spherical Harmonics) and find that we achieve a \textbf{+0.36 dB} improvement, demonstrating that our ray-casting algorithm offers superior rendering quality while avoiding splatting artifacts, as seen in the supplementary videos, which highlights the absence of flickering and splatting artifacts compared to Gaussian Splatting.

\textbf{Basis function:} We then conduct tests to confirm the choice of the Gaussian function. Results show the anisotropic Gaussian outperforms the isotropic one by \textbf{+1.09 dB} on average. Various classical basis functions were also tested, including those with locally native support (such as the Bump function) or globally supported (such as the Inverse Quadratic function). Several showed promising results, notably the Wendland function. However, the anisotropic Gaussian performs best with at least a \textbf{+0.35 dB} difference (see Tab.~\ref{tab:ablation}). This justifies the choice of the anisotropic Gaussian as the basis function.

\textbf{Colorimetric parameters:}
We analyze the impact of different radiance  \(c\) parameterizations: RGB without directional dependence, 16 spherical harmonics (SH), 16 spherical Gaussians (SG), and the combination of 16 SH and SG functions yielding the best results: 9 SH (order 2) and 7 SG. We observe that parameterizations with directional dependence (SH, SG, and SH+SG) significantly improve quality. Then, the parameterization providing the best quantitative results is the combination of SH and SG. This can be justified because SHs are robust for representing low-frequency phenomena, while SGs can adapt to high frequencies.

\section{Conclusions, Limitations and Future Work}
\label{sec:conclusion}
\textbf{Limitations and Future Work}
Although our approach enables high-quality rendering, it has some limitations. Ray casting is computationally intensive, especially with irregularly distributed Gaussians, which results in training times that are not as fast as the current fastest methods, such as Gaussian Splatting. Furthermore, there is still progress in achieving photorealism, for example, by modeling more complex phenomena such as light scattering. Another approach could be to convert our model into a surface model to more easily handle problematic such as scene relighting.

\textbf{Conclusion}
Our approach leverages volume ray casting on Gaussian basis functions associated with Spherical Harmonics/Gaussians colorimetric parameters to optimize scene radiance fields and generate state-of-the-art photorealistic renderings, as quantitative and qualitative results demonstrate. Furthermore, we achieve reasonable training times and fast inference times thanks to our implementation of a slab-by-slab ray casting algorithm using a BVH.

{\small
\bibliographystyle{ieee_fullname}
\bibliography{egbib}

\begin{thebibliography}{10}\itemsep=-1pt

\bibitem{Mip_NeRF}
Jonathan~T. Barron, Ben Mildenhall, Matthew Tancik, Peter Hedman, Ricardo Martin-Brualla, and Pratul~P. Srinivasan.
\newblock Mip-nerf: A multiscale representation for anti-aliasing neural radiance fields.
\newblock In {\em 2021 IEEE/CVF International Conference on Computer Vision (ICCV)}, pages 5835--5844, 2021.

\bibitem{Mip_NeRF_360}
Jonathan~T. Barron, Ben Mildenhall, Dor Verbin, Pratul~P. Srinivasan, and Peter Hedman.
\newblock Mip-nerf 360: Unbounded anti-aliased neural radiance fields.
\newblock In {\em 2022 IEEE/CVF Conference on Computer Vision and Pattern Recognition (CVPR)}, pages 5460--5469, 2022.

\bibitem{zipnerf}
Jonathan~T. Barron, Ben Mildenhall, Dor Verbin, Pratul~P. Srinivasan, and Peter Hedman.
\newblock Zip-nerf: Anti-aliased grid-based neural radiance fields.
\newblock In {\em 2023 IEEE/CVF International Conference on Computer Vision (ICCV)}, pages 19640--19648, 2023.

\bibitem{chandrasekhar2013radiative}
Subrahmanyan Chandrasekhar.
\newblock Radiative transfer.
\newblock {\em Quarterly Journal of the Royal Meteorological Society}, 76(330):498--498, 1950.

\bibitem{chen2024survey3dgaussiansplatting}
Guikun Chen and Wenguan Wang.
\newblock A survey on 3d gaussian splatting, 2024.

\bibitem{chen2023neurbf}
Zhang Chen, Zhong Li, Liangchen Song, Lele Chen, Jingyi Yu, Junsong Yuan, and Yi Xu.
\newblock Neurbf: A neural fields representation with adaptive radial basis functions.
\newblock In {\em Proceedings of the IEEE/CVF International Conference on Computer Vision (ICCV)}, pages 4182--4194, October 2023.

\bibitem{9880067}
Kangle Deng, Andrew Liu, Jun-Yan Zhu, and Deva Ramanan.
\newblock Depth-supervised nerf: Fewer views and faster training for free.
\newblock In {\em 2022 IEEE/CVF Conference on Computer Vision and Pattern Recognition (CVPR)}, pages 12872--12881, 2022.

\bibitem{Plenoxels}
Sara Fridovich-Keil, Alex Yu, Matthew Tancik, Qinhong Chen, Benjamin Recht, and Angjoo Kanazawa.
\newblock Plenoxels: Radiance fields without neural networks.
\newblock In {\em 2022 IEEE/CVF Conference on Computer Vision and Pattern Recognition (CVPR)}, pages 5491--5500, 2022.

\bibitem{Hadwiger2018SparseLeap}
Markus Hadwiger, Ali~K. Al-Awami, Johanna Beyer, Marco Agus, and Hanspeter Pfister.
\newblock Sparseleap: Efficient empty space skipping for large-scale volume rendering.
\newblock {\em IEEE Transactions on Visualization and Computer Graphics}, 24(1):974--983, 2018.

\bibitem{Haines2019}
Eric Haines, Johannes G{\"u}nther, and Tomas Akenine-M{\"o}ller.
\newblock {\em Precision Improvements for Ray/Sphere Intersection}, pages 87--94.
\newblock Apress, Berkeley, CA, 2019.

\bibitem{ji2024mixlightborrowingbestspherical}
Xinlong Ji, Fangneng Zhan, Shijian Lu, Shi-Sheng Huang, and Hua Huang.
\newblock Mixlight: Borrowing the best of both spherical harmonics and gaussian models, 2024.

\bibitem{3D_Gaussian_Splatting}
Bernhard Kerbl, Georgios Kopanas, Thomas Leimkuehler, and George Drettakis.
\newblock 3d gaussian splatting for real-time radiance field rendering.
\newblock {\em ACM Trans. Graph.}, 42(4), jul 2023.

\bibitem{Kingma2014AdamAM}
Diederik Kingma and Jimmy Ba.
\newblock Adam: A method for stochastic optimization.
\newblock In {\em International Conference on Learning Representations (ICLR)}, San Diega, CA, USA, 2015.

\bibitem{Knoll2019}
Aaron Knoll, R.~Keith Morley, Ingo Wald, Nick Leaf, and Peter Messmer.
\newblock {\em Efficient Particle Volume Splatting in a Ray Tracer}, pages 533--541.
\newblock Apress, Berkeley, CA, 2019.

\bibitem{Knoll2014RBFVR}
Aaron Knoll, Ingo Wald, Paul Navratil, Anne Bowen, Khairi Reda, Michael~E. Papka, and Kelly Gaither.
\newblock Rbf volume ray casting on multicore and manycore cpus.
\newblock {\em Computer Graphics Forum}, 33(3):71--80, 2014.

\bibitem{RadTransfinStelAtmo}
{Kubát, J.}
\newblock Radiative transfer in stellar atmospheres.
\newblock {\em EAS Publications Series}, 43:1--18, 2010.

\bibitem{Tetra_NeRF}
J. Kulhanek and T. Sattler.
\newblock Tetra-nerf: Representing neural radiance fields using tetrahedra.
\newblock In {\em 2023 IEEE/CVF International Conference on Computer Vision (ICCV)}, pages 18412--18423, Los Alamitos, CA, USA, oct 2023. IEEE Computer Society.

\bibitem{Perceptual_Quality_Assessment}
H. Liang, T. Wu, P. Hanji, F. Banterle, H. Gao, R. Mantiuk, and C. Öztireli.
\newblock Perceptual quality assessment of nerf and neural view synthesis methods for front-facing views.
\newblock {\em Computer Graphics Forum}, 43(2):e15036, 2024.

\bibitem{MaxOpt}
N. Max.
\newblock Optical models for direct volume rendering.
\newblock {\em IEEE Transactions on Visualization and Computer Graphics}, 1(2):99--108, 1995.

\bibitem{Max2010LocalAG}
Nelson Max and Min Chen.
\newblock {Local and Global Illumination in the Volume Rendering Integral}.
\newblock In Hans Hagen, editor, {\em Scientific Visualization: Advanced Concepts}, volume~1 of {\em Dagstuhl Follow-Ups}, pages 259--274. Schloss Dagstuhl -- Leibniz-Zentrum f{\"u}r Informatik, Dagstuhl, Germany, 2010.

\bibitem{Neural_Radiance_Fields}
Ben Mildenhall, Pratul~P. Srinivasan, Matthew Tancik, Jonathan~T. Barron, Ravi Ramamoorthi, and Ren Ng.
\newblock Nerf: Representing scenes as neural radiance fields for view synthesis.
\newblock In Andrea Vedaldi, Horst Bischof, Thomas Brox, and Jan-Michael Frahm, editors, {\em Computer Vision -- ECCV 2020}, pages 405--421, Cham, 2020. Springer International Publishing.

\bibitem{Instant_NGP}
Thomas M\"{u}ller, Alex Evans, Christoph Schied, and Alexander Keller.
\newblock Instant neural graphics primitives with a multiresolution hash encoding.
\newblock {\em ACM Trans. Graph.}, 41(4), jul 2022.

\bibitem{novak18monte}
Jan Novák, Iliyan Georgiev, Johannes Hanika, and Wojciech Jarosz.
\newblock Monte carlo methods for volumetric light transport simulation.
\newblock {\em Computer Graphics Forum}, 37(2):551--576, 2018.

\bibitem{rbfpower}
J. Park and I.~W. Sandberg.
\newblock Universal approximation using radial-basis-function networks.
\newblock {\em Neural Computation}, 3(2):246--257, 1991.

\bibitem{Parker10OptiX}
Steven~G. Parker, James Bigler, Andreas Dietrich, Heiko Friedrich, Jared Hoberock, David Luebke, David McAllister, Morgan McGuire, Keith Morley, Austin Robison, and Martin Stich.
\newblock Optix: a general purpose ray tracing engine.
\newblock {\em ACM Trans. Graph.}, 29(4), jul 2010.

\bibitem{stopthepop}
Lukas Radl, Michael Steiner, Mathias Parger, Alexander Weinrauch, Bernhard Kerbl, and Markus Steinberger.
\newblock {StopThePop: Sorted Gaussian Splatting for View-Consistent Real-time Rendering}.
\newblock {\em ACM Transactions on Graphics}, 4(43), 2024.

\bibitem{KiloNeRF}
Christian Reiser, Songyou Peng, Yiyi Liao, and Andreas Geiger.
\newblock Kilonerf: Speeding up neural radiance fields with thousands of tiny mlps.
\newblock In {\em 2021 IEEE/CVF International Conference on Computer Vision (ICCV)}, pages 14315--14325, 2021.

\bibitem{schoenberger2016sfm}
Johannes~L. Schönberger and Jan-Michael Frahm.
\newblock Structure-from-motion revisited.
\newblock In {\em 2016 IEEE Conference on Computer Vision and Pattern Recognition (CVPR)}, pages 4104--4113, 2016.

\bibitem{Stamnes_Thomas_Stamnes_2017}
Knut Stamnes, Gary~E. Thomas, and Jakob~J. Stamnes.
\newblock {\em Radiative Transfer in the Atmosphere and Ocean}.
\newblock Cambridge University Press, 2 edition, 2017.

\bibitem{Direct_Voxel_Grid_optimization}
Cheng Sun, Min Sun, and Hwann-Tzong Chen.
\newblock Direct voxel grid optimization: Super-fast convergence for radiance fields reconstruction.
\newblock In {\em 2022 IEEE/CVF Conference on Computer Vision and Pattern Recognition (CVPR)}, pages 5449--5459, 2022.

\bibitem{sun2023pointnerfpp}
Weiwei {Sun}, Eduard {Trulls}, Yang-Che {Tseng}, Sneha {Sambandam}, Gopal {Sharma}, Andrea {Tagliasacchi}, and Kwang {Moo Yi}.
\newblock {PointNeRF++: A multi-scale, point-based Neural Radiance Field}.
\newblock {\em arXiv e-prints}, page arXiv:2312.02362, Dec. 2023.

\bibitem{Point_NeRF}
Qiangeng Xu, Zexiang Xu, Julien Philip, Sai Bi, Zhixin Shu, Kalyan Sunkavalli, and Ulrich Neumann.
\newblock Point-nerf: Point-based neural radiance fields.
\newblock In {\em 2022 IEEE/CVF Conference on Computer Vision and Pattern Recognition (CVPR)}, pages 5428--5438, 2022.

\bibitem{Yu2023MipSplatting}
Zehao Yu, Anpei Chen, Binbin Huang, Torsten Sattler, and Andreas Geiger.
\newblock Mip-splatting: Alias-free 3d gaussian splatting.
\newblock In {\em Proceedings of the IEEE/CVF Conference on Computer Vision and Pattern Recognition (CVPR)}, pages 19447--19456, June 2024.

\bibitem{Differentiable_Point_Based}
Qiang Zhang, Seung-Hwan Baek, Szymon Rusinkiewicz, and Felix Heide.
\newblock Differentiable point-based radiance fields for efficient view synthesis.
\newblock In {\em SIGGRAPH Asia 2022 Conference Papers}, SA '22, New York, NY, USA, 2022. Association for Computing Machinery.

\bibitem{ewa}
M. Zwicker, H. Pfister, J. van Baar, and M. Gross.
\newblock Ewa volume splatting.
\newblock In {\em Proceedings Visualization, 2001. VIS '01.}, pages 29--538, 2001.

\end{thebibliography}
}

\clearpage
\setcounter{section}{0}

\title{Supplementary \\ RayGauss: Volumetric Gaussian-Based Ray Casting \\for Photorealistic Novel View Synthesis}
\author{}
\date{}
\maketitle

\section{Overview}

In this supplementary material, we first provide technical details about our method: an explicit description of the different basis functions tested and their influence on various aspects of the algorithm, details about our OptiX implementation, and a description of optimization details. Next, we address secondary features specific to our approach: the ability to simultaneously cast rays from different viewpoints compared to splatting and the possibilities enabled by the OptiX API\cite{Parker10OptiX}. Finally, we discuss additional tests conducted on the hyperparameters of our algorithm: the sampling step $\Delta t$ and the density threshold $\sigma_\epsilon$. We then compare our method to state-of-the-art approaches attempting to produce an antialiased scene representation. For this purpose, we test our approach by adding brute force supersampling to study its potential for antialiasing.

\section{Description of the different basis functions and their intersection}
We provide more details here on the different tested basis functions and the resulting treatments. Our implementation is flexible and allows easy modification of the basis function used. Such modification impacts the following aspects of the code: construction of the Bounding Volume Hierarchy, intersection program, and weights evaluation for each sample, as explained below.

\subsection{Explicit expressions of the studied basis functions}
As explained in the main article, we limit ourselves to the study of decreasing radial and elliptical basis functions, which can be expressed depending on \( r(\mathbf{x}) = \frac{d_2(\mathbf{x}, \mu)}{R} \) in the radial case or \( r(\mathbf{x}) = d_{M}(\mathbf{x},\mu) \) in the elliptical case. Here, \( R \in \mathbb{R}\) can be interpreted as a shape parameter, \( d_2 \) denotes the Euclidean distance in \( \mathbb{R}^3 \), and \( d_{M}(\mathbf{x},\mu) = \sqrt{(\mathbf{x} - \mathbf{\mu})^T \mathbf{\Sigma}^{-1} (\mathbf{x} - \mathbf{\mu})} \) represents the Mahalanobis distance associated with the covariance matrix \( \Sigma \) and mean position \( \mu \).

Also, we have studied two types of basis functions: those with local, or more precisely compact, support that vanish for \( r > 1 \), and those with global support. We describe below these different functions.

\textbf{Functions with local/compact support:} The studied functions with compact support are as follows:
\begin{itemize}
    \item The Bump function is defined as:
    \begin{equation}
            \phi_{\textit{Bump}}(r) = \begin{cases} 
    e^{1-\frac{1}{1 - r^2}} & \text{if } r < 1 \\
    0 & \text{if } r \geq 1
    \end{cases}
    \end{equation}

    \item Wendland functions denote a class of functions, and here we use one of the most commonly employed:
    \begin{equation}
    \phi_{\textit{Wendland}}(r) = \begin{cases} 
    (1 - r)_{+}^4(4r + 1) & \text{if } r \leq 1 \\
    0 & \text{if } r > 1
    \end{cases}
    \end{equation}
\end{itemize}
It can be noted that the Bump function has been slightly modified compared to its usual expression, as it has been multiplied by $e^1$, so that $\phi_{\textit{Bump}}(0)=1$, which corresponds to the behavior of the other basis functions studied and facilitates experiments.

\textbf{Functions with global support:} The studied functions with global support are as follows:
\begin{itemize}
    \item The inverse multiquadric function:
    \begin{equation}
    \phi_{\textit{Inv\_Multi}}(r) = \frac{1}{\sqrt{1 + r^2}} 
    \end{equation}
    
    \item The inverse quadratic function:
    \begin{equation}
    \phi_{\textit{Inv\_Quad}}(r) = \frac{1}{1 + r^2} 
    \end{equation}
    
    \item The C\(^0\)-Matérn function:
    \begin{equation}
    \phi_{\textit{C0-Matérn}}(r) = \exp(-r) 
    \end{equation}
    
    \item The Gaussian function:
    \begin{equation}
    \phi_{\textit{Gaussian}}(r) = \exp\left(-\frac{r^2}{2}\right) 
    \end{equation}
\end{itemize}

Also, the expression used to evaluate weights for each sample corresponds to one of the expressions given above, depending on the tested case. In addition, as specified in the main article, global support functions are truncated to consider only their value within a domain restricted by the following condition:
\begin{equation}
    \sigma_l(\mathbf{x})_{\text{approx}}=
\begin{cases}
    \sigma_l(\mathbf{x}) & \text{if } \sigma_l(\mathbf{x}) \geq \sigma_{\epsilon} \\
    0 & \text{if } \sigma_l(\mathbf{x}) < \sigma_{\epsilon}
\end{cases}
\label{eq:sigma_approx}
\end{equation}
where:
\begin{equation}
\sigma_l(\mathbf{x}) = \tilde{\sigma_l} \cdot \phi_l(\mathbf{x})
\label{eq:sigma_l}
\end{equation}
with $\phi_l$ the l-th basis function and $\tilde{\sigma_l}$ the associated density parameter.
This allows considering only functions with a non-negligible contribution at a given point by setting a sufficiently low threshold. Moreover, no domain truncation is necessary for compactly supported functions, as they naturally vanish outside a compact domain by their definition.

After modification, the support of these functions corresponds to a solid sphere if the functions are radial or to a solid ellipsoid if they are elliptical. Since elliptical functions yield better results in our case, we will subsequently explain only the case of ellipsoids (which is also a generalization of spheres). Additionally, the OptiX API we use does not natively support intersection with ellipsoids but allows for the definition of a custom intersection program. This works by defining two aspects: the definition of axis-aligned bounding boxes (AABBs) that encompass each primitive and the definition of the custom primitive intersection (ellipsoid in our case). Thus, for a given ray and a given slab assimilated to a segment, we test the intersection of the segment with primitives in the following way: OptiX first finds an intersection with the AABBs, and for the intersected AABBs, it executes the custom program defined by us, calculating the intersection between a segment and an ellipsoid.
Furthermore, the ellipsoid depends on the basis function used. So, as previously mentioned, the basis function influences the construction of the BVH through the provided AABBs and the custom intersection program. We will discuss these two points below. 

 \subsection{Ellipsoid definition for the basis function}
To calculate the intersection of a ray with the support of a basis function, we must first explicitly define its expression. In the case of truncated global functions, by using equations \ref{eq:sigma_approx} and \ref{eq:sigma_l}, we derive that the support of the l-th truncated global function is the set of points satisfying the condition:

\begin{equation}
\phi_l(r(\mathbf{x})) \geq \frac{\sigma_\epsilon}{\tilde{\sigma}_l}
\end{equation}

Also, since all tested global support basis functions are invertible and decreasing, $\phi_l^{-1}$ is decreasing, and:
\begin{equation}
r(\mathbf{x}) \leq \phi_l^{- 1}(\frac{\sigma_\epsilon}{\tilde{\sigma}_l})
\end{equation}
Here, since we are considering the elliptical case, we have:
\begin{equation}
\sqrt{(\mathbf{x} - \mathbf{\mu}_l)^T {\mathbf{\Sigma}_l}^{-1} (\mathbf{x} - \mathbf{\mu}_l)}\leq \phi_l^{- 1}(\frac{\sigma_\epsilon}{\tilde{\sigma}_l})
\end{equation}
And thus (by the growth of the square function on $\mathbb{R}^{+}$):
\begin{equation}
(\mathbf{x} - \mathbf{\mu}_l)^T {\mathbf{\Sigma}_l}^{-1} (\mathbf{x} - \mathbf{\mu}_l)\leq \left(\phi_l^{- 1}(\frac{\sigma_\epsilon}{\tilde{\sigma}_l})\right)^2
\label{eq:ellips_global}
\end{equation}
where we can recognize the equation of an ellipsoid.

The expression is simpler in the case of locally/compactly supported functions. Indeed, the tested functions vanish starting from \( r=1 \). Therefore, we can deduce the equation of the ellipsoid in this case:

\begin{equation}
(\mathbf{x} - \mathbf{\mu}_l)^T {\mathbf{\Sigma}_l}^{-1} (\mathbf{x} - \mathbf{\mu}_l)\leq 1
\label{eq:ellips_local}
\end{equation}
\subsection{Tightest axis-aligned bounding box definition}
The OptiX API optimizes Ray-AABB intersection. Also, our goal here is to build the BVH using the tightest axis-aligned bounding boxes enclosing previous ellipsoids to avoid unnecessary intersection calculations. In this elliptical case, the resulting axis-aligned bounding box (AABB) has a slightly complex expression. It is bounded by the following coordinates:
\begin{equation}
\begin{aligned}
x &= \mu_x \pm \sqrt{{\tilde{s}_x}^2\mathbf{R}_{1,1}^2 + {\tilde{s}_y}^2\mathbf{R}_{1,2}^2 + {\tilde{s}_z}^2\mathbf{R}_{1,3}^2}, \\
y &= \mu_y \pm \sqrt{{\tilde{s}_x}^2\mathbf{R}_{2,1}^2 + {\tilde{s}_y}^2\mathbf{R}_{2,2}^2 + {\tilde{s}_z}^2\mathbf{R}_{2,3}^2}, \\
z &= \mu_z \pm \sqrt{{\tilde{s}_x}^2\mathbf{R}_{3,1}^2 + {\tilde{s}_y}^2\mathbf{R}_{3,2}^2 + {\tilde{s}_z}^2\mathbf{R}_{3,3}^2}.
\end{aligned}
\end{equation}
where $\mathbf{R}_{i,j}$ are the coordinates of the rotation matrix $\mathbf{R}$ associated with the ellipsoid as described in the main article. 
Moreover, $\tilde{\mathbf{s}} = \mathbf{s} \cdot \phi_l^{-1}\left(\frac{\sigma_\epsilon}{\tilde{\sigma}_l}\right)$ in the case of truncated global functions and $\tilde{\mathbf{s}} = \mathbf{s}$ in the case of compactly supported basis functions. Here, $\mathbf{s}$ refers to the diagonal of the scale matrix $\mathbf{S}$ defined in the main paper.

In the radial case, this is straightforward as we take  the cube of side $2R\phi_l^{-1}\left(\frac{\sigma_\epsilon}{\tilde{\sigma}_l}\right)$ centered on $\mu$ in the case of truncated global functions and the cube of side $2R$ centered on $\mu$, in the case of compactly supported basis functions.

\subsection{Custom intersection definition}
To implement the intersection, we use the ellipsoid equations \ref{eq:ellips_global} and \ref{eq:ellips_local}, depending on the basis function, and calculate the intersection of the current segment with this ellipsoid. In particular, we leverage the optimized Ray-Sphere intersection introduced in \cite{Haines2019} and adapted for the case of ellipsoids.

\section{Optix Raycasting}
Here we describe the details of our RayCasting algorithm implementation using the OptiX API. This API works by allowing the definition of several custom programs that define the behavior of the rendering pipeline. The different programs of interest in our case are as follows:
\begin{itemize}
    \item Ray Generation: This program is called first and is executed in parallel for each pixel, launching rays into the BVH.
    \item Intersection: Defines the ray-primitive intersection with our custom primitive, an ellipsoid in our case.
    \item Any-Hit: This program is called when the Intersection program finds a new intersection along the ray, allowing custom processing of the intersected primitives.
\end{itemize}

From these three programs, we can define our ray-casting algorithm. In particular, the Ray Generation program calculates the origin and direction of the ray associated with a given pixel, then a slab size corresponding to a multiple of $\Delta t$ is fixed. This slab size is the same for all rays to maintain inter-ray coherence. In practice, in most cases, we treat 8 samples per slab. Next, if a given ray intersects the axis-aligned bounding box associated with the set of primitives, then this ray intersects the scene. So we start probing the space traversed by the ray, proceeding slab by slab. For a given segment on the ray, we launch the traversal of the BVH. The intersection program computes the intersection of the segment with ellipsoids in the scene. The any-hit program collects the primitives intersecting the segment by storing their index in a large pre-allocated buffer. In practice, the buffer size is set to store at most 512 or 1024 primitives depending on the scene, thus allowing us to have more than enough primitives per buffer (between 512 and 1024 for 8 samples). This choice was made to ensure fast code execution. Once the primitives contributing locally are collected, we can calculate the value of $\sigma$ and $c$ for each of the samples in the slab, then accumulate them in the form of an intermediate color and transmittance of the ray. We can then use the early termination strategy, which ends the calculation when the current transmittance becomes lower than a threshold: \( T < T_{\epsilon}\). When the threshold is low, this strategy allows us to disregard the samples whose contribution will be negligible compared to the overall color of the ray. If early termination isn't applied and we are still within the bounds of the scene, we can then move to the next slab and repeat the same process. We provide the ray generation and any-hit programs for performing ray casting in algorithm \ref{alg:any_hit} and \ref{alg:ray_gen}. The intersection program is not described in detail as it corresponds to the ellipsoid-segment intersection.

\begin{algorithm}[h] 
\small
\caption{Any-Hit Program}
\textbf{Input:}  \(n_p\): number of intersected primitives, $n_{max}$: maximum number of primitives, hitBuffer: buffer storing primitive indices, $i_R$: index of the current ray \\
\textbf{Output:}  \(n_p\), hitBuffer
\begin{algorithmic}[1]
    \State $i_P \gets \text{optixGetPrimitiveIndex()}$ \Comment{Current primitive index}    
    \State $\text{hitBuffer}[i_R \times n_{max} + n_p] \gets i_P$
    \State $n_p \gets n_p+1$ 
    \If{$n_p \geq n_{max}$}
        \State \text{optixTerminateRay()} \Comment{Terminate if max. primitives}
    \EndIf
    \State \text{optixIgnoreIntersection()} \Comment{Continue Traversal}

\end{algorithmic}
\label{alg:any_hit}
\end{algorithm}

\algnewcommand{\LineComment}[1]{\State \(\triangleright\) #1}

\begin{algorithm}[h]
\small
\caption{Ray Generation Program}
\textbf{Input:}  $i_R$: index of the current ray, $bbox_{min}$: minimum bounds of the bounding box, $bbox_{max}$: maximum bounds of the bounding box, $\Delta t$: step size, $B$: size of the buffer, $T_{\epsilon}$: transmittance threshold, hitBuffer: buffer storing primitive indices, P: global parameters (primitive parameters and ray colors) \\
\textbf{Output:}  P: update parameters
\begin{algorithmic}[1]
    \State $o, d \gets \text{ComputeRay}(i_R)$ \Comment{Ray origin, direction}
    \State $t_0, t_1 \gets \text{IntersectBBox}(o,d,bbox_{min},bbox_{max})$
    \State $\Delta S\gets \Delta t \times B$ \Comment{Slab size}
    \State $T \gets 1.0$ \Comment{Ray transmittance}
    \State $C_R \gets (0.0,0.0,0.0)$ \Comment{Ray color}

    \If{$t_0 < t_1$} \Comment{Check if ray intersects bounding box}
        \State $t_S \gets t_0$ \Comment{Current slab distance along the ray}
        \While{$t_S < t_1$ \textbf{and} $T > T_{\epsilon}$}
            \State $n_p \gets 0$ \Comment{Number of primitives}
            \State $t_{min\_S} \gets \text{max}(t_0, t_S)$
            \State $t_{max\_S} \gets \text{min}(t_1, t_S + \Delta S)$
            \LineComment{Collect the intersected primitives}
                \State $\text{Traversal}(\text{hitBuffer}, o, d, t_{min\_S}, t_{max\_S}, n_p)$
                \If{$n_p == 0$}
                    \State $t_S \gets t_S + \Delta S$
                    \State \textbf{continue}
                \EndIf
                \State $\text{densityBuffer} \gets (0.0)^{B}$
                \State $\text{colorBuffer} \gets (0.0,0.0,0.0)^{B}$
                \LineComment{Update ray color and density}
                \State $\text{UpdateRay}(i_R, n_p, \Delta t, t_S, o, d, \text{densityBuffer},$
                \Statex \hspace{10.5em} $\text{colorBuffer}, C_R, T, \text{P})$
            \State $t_S \gets t_S + \Delta S$
        \EndWhile
    \EndIf
    \State $\text{P.ray\_colors}[i_R]\gets C_R$
\end{algorithmic}
\label{alg:ray_gen}
\end{algorithm}

\section{Optimization details}
This section provides further details on the optimization parameters used in our experiments. We use the Adam gradient descent optimization algorithm \cite{Kingma2014AdamAM}. We recall that the parameters optimized by our approaches are as follows: \( P = \{ (\tilde{\sigma}_l, \tilde{c}_l, \mu_l, \mathbf{q}_l, \mathbf{s}_l) \mid l = 1, \ldots, N \} \)
where $\tilde{\sigma}_l$ is the density parameter of the $l$-th primitive, $\tilde{c}_l$ summarizes the colorimetric parameters: lobe  \(\lambda_{l,j}\) , lobe direction \(\mathbf{p}_{l,j}\), coefficients \(k_{l,j}\), for the $j-th$ Spherical Gaussian and coefficients  \(\tilde{c}_{l,jm}\) for Spherical Harmonic of degree \(j\) and order \(m\), while $\mu_l$, $\mathbf{q}_l$, and $\mathbf{s}_l$ are the mean position, quaternion, and scale parameter used to evaluate the basis function of the $l$-th primitive.  The initial parameters are obtained from a point cloud created by Structure-From-Motion methods such as Colmap\cite{schoenberger2016sfm}. 

We will now describe the main parameters of our optimization, starting with the learning rates. 
In the context of evaluations on the Blender dataset, we employed a learning rate for density with exponential decay, starting at \(5 \times 10^{-1}\) and decreasing to \(3 \times 10^{-2}\) over 30,000 iterations. The learning rates for other parameters were set as follows: \( 1.0 \times 10^{-3} \) for constant color parameters (RGB), \( 2.6 \times 10^{-4} \) for Spherical Harmonics coefficients, \( 3.6 \times 10^{-4} \) for Spherical Gaussians coefficients, \( 4.5 \times 10^{-2} \) for lobe sharpness, \( 3.0 \times 10^{-3} \) for lobe direction, \( 1.2 \times 10^{-2} \) for scale parameters, \( 2.2 \times 10^{-4} \) for quaternions, and a learning rate with exponential decay starting from \( 1.7 \times 10^{-5} \) and reaching \( 1.0 \times 10^{-6} \) after 30,000 iterations for mean positions. We optimize the scene using a white background to place ourselves in evaluation conditions like other state-of-the-art methods. In the case of the Mip-NeRF360 dataset, we use similar learning rates except for the density, for which we use a learning rate with exponential decay starting from  \(5 \times 10^{-1}\) and reaching  \(1 \times 10^{-4}\) in 30,000 iterations. This allows us to accelerate the beginning of the optimization, particularly for outdoor scenes with a large number of Gaussians. Furthermore, we train with a black background on this dataset to place ourselves in evaluation conditions similar to Gaussian Splatting and current state-of-the-art methods \cite{3D_Gaussian_Splatting}\cite{Yu2023MipSplatting}. 

More generally, we optimize the scene by backpropagation, with the optimization taking 30,000 iterations (1 image per iteration). Quaternions are renormalized after each optimization step to maintain a valid rotation matrix representation. The process of Adaptive Gaussian Control is similar to that used in 3D Gaussian Splatting \cite{3D_Gaussian_Splatting}, periodically increasing, every \(i_{densify}\) iterations, the number of Gaussians in the scene using a heuristic based on the condition \(\nabla L_{\mu_l} > \nabla L_{\epsilon}\), where \(\nabla L_{\mu_l}\) denotes the gradient on the mean position of the basis functions and \(\nabla L_{\epsilon}\) the threshold from which densification is applied. In our case, we densify every 500 iterations for Mip-NeRF360 and 300 iterations for Blender dataset, starting from the 500th iteration up to the 15,000th iteration, and we set \(\nabla L_{\epsilon} = 0.00002\) for the Mip-NeRF 360 dataset and \(\nabla L_{\epsilon} = 0.00004\) for the Blender dataset. Furthermore, we remove primitives whose density parameter $\tilde{\sigma}$ is below a threshold, set to 0.1 for our experiments on the Blender dataset and 0.01 on the Mip-NeRF 360 dataset. Finally, as described in the main paper, we gradually unlock the colorimetric parameters from the lowest frequency representation to the potentially higher frequency one: harmonics of degree 0, 1, 2, and finally, 7 spherical Gaussians, resulting in 16 functions to represent the color of a primitive. The unlocking of these parameters occurs every 1000 iterations. We can specify here that the spherical harmonic and spherical gaussian coefficients, as well as the lobe sharpness, are initialized to zero, while the lobe axes are randomly initialized on the unit sphere and constrained to it during optimization. Also, one can refer to Algorithm \ref{alg:optim} to gain a broader view of the optimization process.

\begin{algorithm}[h] 
\small
\caption{RayGauss Scene Optimization}
\textbf{Input:}  \( (I_i)_{i=1}^{N} \):$N$ training images, $i_{max}$ maximum number of iterations, $\nabla L_{\epsilon}$ gradient threshold for densification, $\sigma_\epsilon$ density threshold for prunning\\
\textbf{Output:} optimized primitive parameters $P$ 
\begin{algorithmic}[1]
\State \( (V_i)_{i=1}^{N}, \tilde{P} \gets \text{SFM}( (I_i)_{i=1}^{N} ) \) \Comment{Camera, Sparse Point Cloud}
\State \( P \gets \text{InitPrim}(\tilde{P}) \) \Comment{Initialize primitive parameters}
\State \( T \gets \text{InitBVH}(P)\)\Comment{Bounding Volume Hierarchy}
\State \( i \gets 0 \)
\While{\( i < i_{max} \)}
    \State \(V, I \gets \text{SampleTrainView()} \)
    \State \(\hat{I} \gets \text{RayCast}(V, P, T) \)
    \State \(L \gets \text{Loss}(\hat{I},I)\)
    \State \(P\gets \text{AdamOptim}(\nabla L)\)
    \For{\( (\tilde{\sigma}_l, \tilde{c}_l, \mu_l, \mathbf{q}_l, \mathbf{s}_l) \) in \( P \)}
        \If{\( \text{IsAdaptControlIter($i$)} \)}
            \If{\( \nabla L_{\mu_l}>\nabla L_{\epsilon} \)}
            \State  \(P\gets \text{CloneSplit}( (\tilde{\sigma}_l, \tilde{c}_l, \mu_l, \mathbf{q}_l, \mathbf{s}_l))\)
            \EndIf 
            \If{\(\tilde{\sigma}_l<\sigma_{\epsilon}\)}
                \State \text{RemovePrimitive()}
            \EndIf
        \EndIf
        \If{\( \text{IsUnlockIter($i$)} \)}
            \State \(\tilde{c}_l \gets \text{UnlockColorSHSG}(i)\)
        \EndIf
    \EndFor
    \State \(T \gets \text{UpdateBVH}(T,P)\)
\EndWhile

\end{algorithmic}
\label{alg:optim}
\end{algorithm}

Furthermore, it should be noted that the number of parameters in our representation is \( 87N_{P} \), where \( N_{P} \) is the number of primitives. Specifically, we have the following parameter counts: 3 for the mean position \( \mu \), 3 for the scale parameter \( \mathbf{s} \), 4 for the quaternions \( \mathbf{q} \), 1 for the density parameter \( \tilde{\sigma} \), 27 for spherical harmonic coefficients, 21 for spherical Gaussian coefficients, 7 for lobe  parameters, and 21 for lobe direction parameters. Moreover, as mentioned earlier, our current implementation requires allocating a buffer of size \( N_{max,P} \times N_{ray} \), where \( N_{max,P}\) is the maximum number of primitives per ray per slab allowed, and \(N_{ray} \) the number of rays launched. Therefore, the limiting factor of our implementation in terms of memory mainly depends on the allocation of these data.

\section{Uncorrelated ray casting}
One of the advantages of ray casting compared to splatting lies in the fact that multiple independent rays from different cameras can be rendered simultaneously. In ray casting, which is an image-order rendering method, rays are treated independently and can originate from various viewpoints. In contrast, splatting, which is an object-order method, projects primitives onto the image plane and then sorts them. This primitive processing benefits splatting when rendering an entire image, as it allows for quickly computing the color of each pixel by summing the contributions of primitives projected onto it. However, this approach loses its advantage when considering the color of a single ray, for instance. Thus, in practice, several applications can be considered based on this observation: training is done iteratively on individual images in 3D Gaussian Splatting, whereas our training can easily use batches of rays from different images. Furthermore, supervision may require casting independent rays if the supervision data is sparse, for instance, if we want to supervise ray depth using a point cloud representing the surface as supervision data \cite{9880067}. In this case, our approach is more suitable than the splatting algorithm because it can natively handle uncorrelated rays from different viewpoints. Another application would be adaptive supersampling, which consists of successively casting rays in the image plane to reduce rendering artifacts by focusing on the most challenging regions. This type of approach is more suited to ray casting as improving the rendering quality may only require casting a few additional rays. These last two applications are beyond the scope of this article.
Additionally, tests were conducted by training with batches of rays. However, training in batches does not allow for the use of supervision functions such as structural similarity (SSIM), and in practice, we obtained poorer results compared to training image by image. However, conducting more experiments to explore this aspect further would be interesting.

\section{OptiX API applications}
Our method is supported by Nvidia OptiX, a framework designed initially for GPU ray tracing. Its flexible API allows for efficiently combining different types of primitives, associated intersections, and rendering algorithms. Consequently, our approach has the potential to be combined with more traditional rendering methods using standard primitives, such as meshes rendered by classic ray tracing, through the API used. Thus, our approach could be integrated into complex environments mixing different types of primitives that can be rendered using the single OptiX API.

\section{Analysis of $\Delta t$ and $\sigma_{\epsilon}$}

In this section, we present a study of the influence of two parameters on the final rendering quality, training and rendering times:
\begin{itemize}
    \item $\Delta t$ is the distance between two samples along a ray
    \item $\sigma_{\epsilon}$ corresponds both to the density under which a Gaussian is removed but also to the limits of the Gaussians for the calculation of intersections 
\end{itemize}

Tab.~\ref{tab:study_delta_t} studies the influence of the parameter $\Delta t$ on the Blender dataset. The results come from training on the Blender Dataset with PSNR averaged over the 8 scenes of this synthetic dataset. The gray line corresponds to the choice of the parameter $\Delta t$ for all other experiments (main article and supplementary). We observe that increasing the space $\Delta t$ makes it possible to speed up training but also rendering times at the cost of a reduction in graphic quality.

\begin{table}[h]
  \centering
  \resizebox{\columnwidth}{!}{%
  \begin{tabular}{l | c c c }
    $\Delta t$ & \textbf{PSNR ↑} & Training Time & Rendering Time \\
    \hline
    0.00125 & 34.53 & 46 min & 17.4 FPS \\
    \cellcolor{gray!20}0.0025 & \cellcolor{gray!20}34.53 & \cellcolor{gray!20}32 min & \cellcolor{gray!20}25.8 FPS \\
    0.005 & 34.52 & 24 min & 42.8 FPS \\
    0.01 & 33.90 & 20 min & 50.1 FPS \\
  \end{tabular}
  }
  \caption{\textbf{Study of the influence of the parameter $\Delta t$ on the Blender dataset}. $\Delta t$ is the distance between two samples along a ray. Values are averaged per scene. In gray, the parameter used in other experiments. Rendering time is for 800x800 pixels image.}
  \label{tab:study_delta_t}
\end{table}

Tab.~\ref{tab:study_sigma_epsilon} studies the influence of the parameter $\sigma_{\epsilon}$ on the Blender dataset. The results come from training on the Blender Dataset with PSNR averaged over the 8 scenes of this synthetic dataset. We can see that increasing $\sigma_{\epsilon}$ speeds up the training and rendering times of the method. On the contrary, by decreasing $\sigma_{\epsilon}$, we increase the size of the Gaussians when calculating the intersections with the rays, which increases the training and rendering times, while improving the quality of rendering. The gray line also corresponds to the choice of the parameter $\sigma_{\epsilon}$ for all other experiments on Blender dataset (main article and supplementary), for Mip-NeRF360, we use $\sigma_{\epsilon}=0.01$.

\begin{table}[h]
  \centering
  \resizebox{\columnwidth}{!}{%
  \begin{tabular}{l | c c c }
   $\sigma_{\epsilon}$ & \textbf{PSNR ↑} & Training Time per scene & Rendering Time \\
    \hline
     1.0 & 34.52 & 26 min & 29.4 FPS\\
      \cellcolor{gray!20}0.1 & \cellcolor{gray!20}34.53 & \cellcolor{gray!20}32 min & \cellcolor{gray!20}25.8 FPS \\
     0.01 & 34.54 & 35 min & 23.5 FPS\\
  \end{tabular}
  }
  \caption{\textbf{Study of the influence of the parameter $\sigma_{\epsilon}$ on the Blender dataset}. $\sigma_{\epsilon}$ determines the limit of Gaussians for calculating intersections. Values are average per scene. In gray, the parameter used in other experiments. Rendering time is for 800x800 pixels image.}
  \label{tab:study_sigma_epsilon}
\end{table}

\section{RayGauss and Anti-aliasing}

Like NeRF~\cite{Neural_Radiance_Fields}, Instant-NGP~\cite{Instant_NGP} and Gaussian Splatting~\cite{3D_Gaussian_Splatting}, RayGauss is a method that does not have an anti-aliasing mechanism, unlike the Mip-NeRF~\cite{Mip_NeRF} and Mip-Splatting~\cite{Yu2023MipSplatting} methods. 

To study the level of aliasing of RayGauss, we followed the protocol defined by Mip-Splatting~\cite{Yu2023MipSplatting} Single-Scale vs Multi-Scale with the Blender dataset, which consists of training the methods with full resolution images (800x800 pixels) then testing the rendered at different resolutions (1, $\sfrac{1}{2}$, $\sfrac{1}{4}$, $\sfrac{1}{8}$) to mimic zoom-out effects. We can see the results with the PSNR metric on Tab.~\ref{tab:blender_st_mt}. RayGauss manages to maintain good rendering quality on all scales compared to Gaussian Splatting (due to rasterization and dilation of Gaussians in 2D) and remains competitive compared to methods with anti-aliasing (Mip-NeRF~\cite{Mip_NeRF} and Mip-Splatting~\cite{Yu2023MipSplatting}).

\begin{table}[h!]
  \centering
  \resizebox{\columnwidth}{!}{%
  \begin{tabular}{l | c c c c | c }
    & Full Res. & $\sfrac{1}{2}$ Res. & $\sfrac{1}{4}$ Res. & $\sfrac{1}{8}$ Res. & \textbf{Avg.}\\
    \hline
Gaussian Splatting~\cite{Yu2023MipSplatting} & \cellcolor{yellow!40}33.33 & 26.95 & 21.38 & 17.69 & 24.84 \\
NeRF~\cite{Neural_Radiance_Fields} & 31.48 & 32.43 & \cellcolor{yellow!40}30.29 & \cellcolor{yellow!40}26.70 & 30.23 \\
Instant-NGP~\cite{Instant_NGP} & 33.09 & 33.00 & 29.84 & 26.33 & 30.57 \\
MipNeRF~\cite{Mip_NeRF} & 33.08 & 33.31 & \cellcolor{orange!40}30.91 & \cellcolor{orange!40}27.97 & \cellcolor{orange!40}31.31 \\
Mip-Splatting~\cite{Yu2023MipSplatting} & \cellcolor{orange!40}33.36 & \cellcolor{red!40}34.00 & \cellcolor{red!40}31.85 & \cellcolor{red!40}28.67 & \cellcolor{red!40}31.97 \\
\hline
RayGauss (ours) & \cellcolor{red!40}34.53 & \cellcolor{orange!40}33,90 & 30,01 & 26,36 & \cellcolor{yellow!40}31.20 \\
  \end{tabular}
  }
  \caption{\textbf{PSNR score for Single-scale Training and Multi-scale Testing on the Blender dataset.} All methods are trained on full-resolution images (800x800 pixels)
and evaluated at four different resolutions (800x800, 400x400, 200x200 and 100x100 pixels), lower resolutions simulating zoom-out effects.}
  \label{tab:blender_st_mt}
\end{table}

A brute-force anti-aliasing method consists in multiplying the number of rays per pixel. The basic RayGauss method launches a single ray through the center of each pixel for training and rendering. We studied the effect of casting 4 rays per pixel for training and rendering, a variant called RayGauss4x. Training and rendering times are approximately 3 times longer than the basic RayGauss method. We then calculated the PSNR scores on Blender with several scales (Tab.~\ref{tab:blender_st_mt_4x}) and compared it with the Mip-Splatting method. To be fair, we also increased the rasterization resolution of Mip-splatting by 4 at each scale (during training and rendering), a variant called Mip-Splatting4x. In this configuration, RayGauss4x is superior to Mip-Splatting4x on almost all scales on the Blender dataset.

\begin{table}[h!]
  \centering
  \resizebox{\columnwidth}{!}{%
  \begin{tabular}{l | c c c c | c }
    & Full Res. & $\sfrac{1}{2}$ Res. & $\sfrac{1}{4}$ Res. & $\sfrac{1}{8}$ Res. & \textbf{Avg.}\\
    \hline
Mip-Splatting4x~\cite{Yu2023MipSplatting}* & \cellcolor{yellow!40}33.51 & \cellcolor{yellow!40}35.23 & \cellcolor{yellow!40}35.71 & \cellcolor{red!40}33.92 & \cellcolor{yellow!40}34.59 \\
\hline
RayGauss4x (ours) & \cellcolor{red!40}34.60 & \cellcolor{red!40}36.43 & \cellcolor{red!40}36.02 & \cellcolor{yellow!40}33.11 & \cellcolor{red!40}35.04 \\
  \end{tabular}
  }
  \caption{\textbf{PSNR score for Single-scale Training and Multi-scale Testing on the Blender dataset with 4x Super Sampling.} Methods are trained on full-resolution images (800x800 pixels) with 4x supersampling
and evaluated at four different resolutions (800x800, 400x400, 200x200 and 100x100 pixels)  with 4x supersampling, lower resolutions simulating zoom-out effects.}
  \label{tab:blender_st_mt_4x}
\end{table}

\section{Detailed results}

Tab.~\ref{tab:detailed_blender} and Tab.~\ref{tab:detailed_mipnerf360} show the detailed results of the main paper with metrics PSNR, SSIM, and LPIPS on the Blender and Mip-NeRF 360 datasets. Some methods have no available code, so we were not able to report information about SSIM and LPIPS (for example, for PointNeRF++).

All methods with an * have been retrained using the available code:
\begin{itemize}
 \item NeRF and Instant-NGP with their respective model using the nefstudio framework v1.1.3: \url{https://github.com/nerfstudio-project/nerfstudio}
 \item Gaussian Splatting: \url{https://github.com/graphdeco-inria/gaussian-splatting}
 \item Mip-Splatting: \url{https://github.com/autonomousvision/mip-splatting}
\end{itemize}
 
\begin{table*}[h]
  \centering
  \resizebox{0.8\linewidth}{!}{%
  \begin{tabular}{l | c c c c c c c c | c }
   & \multicolumn{8}{c}{\textbf{PSNR ↑}} \\
    & Chair & Drums & Ficus & Hotdog & Lego & Materials & Mic & Ship & \textbf{Avg.}\\
    \hline
NeRF~\cite{Neural_Radiance_Fields} & 34.17 & 25.08 & 30.39 & 36.82 & 33.31 & 30.03 & 34.78 & 29.30 & 31.74 \\
Zip-NeRF~\cite{zipnerf} & 34.84 & 25.84 & 33.90 & 37.14 & 34.84 & \cellcolor{orange!40}31.66 & 35.15 & \cellcolor{yellow!40}31.38 & 33.10 \\
Instant-NGP~\cite{Instant_NGP} & 35.00 & 26.02 & 33.51 & 37.40 & 36.39 & 29.78 & 36.22 & 31.10 & 33.18 \\ 
Mip-NeRF360~\cite{Mip_NeRF_360} & 35.65 & 25.60 & 33.19 & 37.71 & 36.10 & 29.90 & \cellcolor{yellow!40}36.52 & 31.26 & 33.24 \\
Point-NeRF~\cite{Point_NeRF} & 35.40 & 26.06 & \cellcolor{red!40}36.13 & 37.30 & 35.04 & 29.61 & 35.95 & 30.97 & 33.30 \\
Gaussian Splatting~\cite{3D_Gaussian_Splatting}* & 35.85 & 26.22 & 35.00 & 37.81 & 35.87 & 30.00 & 35.40 & 30.95 & 33.39 \\
Mip-Splatting~\cite{Yu2023MipSplatting}* & 36.03 & \cellcolor{yellow!40}26.29 & \cellcolor{orange!40}35.33 & \cellcolor{yellow!40}37.98 & 36.03 & 30.29 & 35.63 & 30.50 & 33.51 \\
PointNeRF++~\cite{sun2023pointnerfpp} & \cellcolor{yellow!40}36.32 & 26.11 & 34.43 & 37.45 & \cellcolor{yellow!40}36.75 & 30.32 & \cellcolor{orange!40}36.85 & 31.34 & \cellcolor{yellow!40}33.70 \\
NeuRBF~\cite{chen2023neurbf}*& \cellcolor{orange!40}36.54 & \cellcolor{orange!40}26.38 & 35.01 & \cellcolor{red!40}38.44 & \cellcolor{red!40}37.35 & \cellcolor{red!40}34.12 & 36.16 & \cellcolor{orange!40}31.73 & \cellcolor{orange!40}34.47 \\
    \hline
RayGauss (ours) & \cellcolor{red!40}37.20 & \cellcolor{red!40}27.14 & 
\cellcolor{yellow!40}35.11 & \cellcolor{orange!40}38.30 & \cellcolor{orange!40}37.10 & \cellcolor{yellow!40}31.36 & \cellcolor{red!40}38.11 & \cellcolor{red!40}31.95 & \cellcolor{red!40}34.53\\
  \end{tabular}
  }
  
\bigskip
   
  \resizebox{0.8\linewidth}{!}{%
  \begin{tabular}{l | c c c c c c c c | c }
   & \multicolumn{8}{c}{\textbf{SSIM ↑}} \\
    & Chair & Drums & Ficus & Hotdog & Lego & Materials & Mic & Ship & \textbf{Avg.}\\
    \hline
NeRF~\cite{Neural_Radiance_Fields} & 0.975 & 0.925 & 0.967 & 0.979 & 0.968 & 0.953 & 0.987 & 0.869 & 0.953 \\
Zip-NeRF~\cite{zipnerf} & 0.983 & 0.944 & \cellcolor{yellow!40}0.985 & 0.984 & 0.980 & \cellcolor{orange!40}0.969 & \cellcolor{yellow!40}0.991 & \cellcolor{red!40}0.929 & \cellcolor{orange!40}0.971 \\
Instant-NGP~\cite{Instant_NGP} & - & - & - & - & - & - & - & - & - \\
Mip-NeRF360~\cite{Mip_NeRF_360} & 0.983 & 0.931 & 0.979 & 0.982 & 0.980 & 0.949 & \cellcolor{yellow!40}0.991 & 0.893 & 0.961 \\
Point-NeRF~\cite{Point_NeRF} & \cellcolor{yellow!40}0.984 & 0.935 &  \cellcolor{orange!40}0.987 & 0.982 & 0.978 & 0.948 & 0.990 & 0.892 & 0.962\\
Gaussian Splatting~\cite{3D_Gaussian_Splatting}* &  \cellcolor{orange!40}0.988 & \cellcolor{yellow!40}0.955 & \cellcolor{red!40}0.988 &  \cellcolor{yellow!40}0.986 & \cellcolor{yellow!40}0.983 & 0.960 &  \cellcolor{orange!40}0.992 & 0.893 & 0.968 \\
Mip-Splatting~\cite{Yu2023MipSplatting}* &  \cellcolor{orange!40}0.988 &  \cellcolor{orange!40}0.956 & \cellcolor{red!40}0.988 & \cellcolor{orange!40}0.987 &  \cellcolor{orange!40}0.984 &  \cellcolor{yellow!40}0.962 &  \cellcolor{orange!40}0.992 & 0.900 & \cellcolor{yellow!40}0.970 \\
Point-NeRF++~\cite{sun2023pointnerfpp} & - & - & - & - & - & - & - & - & - \\
NeuRBF\cite{chen2023neurbf}* & \cellcolor{orange!40}0.988 & 0.944 & \cellcolor{orange!40}0.987 & \cellcolor{orange!40}0.987 & \cellcolor{red!40}0.986 & \cellcolor{red!40}0.979 & \cellcolor{orange!40}0.992 & \cellcolor{orange!40}0.925 & \cellcolor{red!40}0.974 \\
\hline
RayGauss (ours) & \cellcolor{red!40}0.990 & \cellcolor{red!40}0.960 & \cellcolor{red!40}0.988 & \cellcolor{red!40}0.988 & \cellcolor{red!40}0.986 & \cellcolor{orange!40}0.969 & \cellcolor{red!40}0.995 &  \cellcolor{yellow!40}0.914 & \cellcolor{red!40}0.974 \\
  \end{tabular}
  }

\bigskip

  \resizebox{0.8\linewidth}{!}{%
  \begin{tabular}{l | c c c c c c c c | c }
   & \multicolumn{8}{c}{\textbf{LPIPS ↓}} \\
    & Chair & Drums & Ficus & Hotdog & Lego & Materials & Mic & Ship & \textbf{Avg.}\\
    \hline
NeRF~\cite{Neural_Radiance_Fields} & 0.026 & 0.071 & 0.032 & 0.030 & 0.031 & 0.047 & 0.012 & 0.150 & 0.050 \\
Zip-NeRF~\cite{zipnerf} & 0.017 & \cellcolor{yellow!40}0.050 & \cellcolor{orange!40}0.015 & 0.020 & 0.019 & \cellcolor{orange!40}0.032 & 0.007 & \cellcolor{orange!40}0.091 & \cellcolor{yellow!40}0.031 \\
Instant-NGP~\cite{Instant_NGP} & - & - & - & - & - & - & - & - & - \\
Mip-NeRF360~\cite{Mip_NeRF_360} & 0.018 & 0.069 & 0.022 & 0.024 & \cellcolor{yellow!40}0.018 & 0.053 & 0.011 & 0.119 & 0.042 \\
Point-NeRF~\cite{Point_NeRF} & 0.023 & 0.078 & 0.022 & 0.037 & 0.024 & 0.072 & 0.014 & 0.124 & 0.049 \\
Gaussian Splatting~\cite{3D_Gaussian_Splatting}* & \cellcolor{orange!40}0.011 & \cellcolor{orange!40}0.037 & \cellcolor{red!40}0.011 & \cellcolor{orange!40}0.017 & \cellcolor{orange!40}0.015 & 0.034 & \cellcolor{yellow!40}0.006 & 0.118 & \cellcolor{yellow!40}0.031\\
Mip-Splatting~\cite{Yu2023MipSplatting}* & \cellcolor{yellow!40}0.012 & \cellcolor{orange!40}0.037 & \cellcolor{red!40}0.011 & \cellcolor{yellow!40}0.018 & \cellcolor{orange!40}0.015 & \cellcolor{yellow!40}0.033 & \cellcolor{orange!40}0.005 & \cellcolor{yellow!40}0.107 & \cellcolor{orange!40}0.030 \\
Point-NeRF++~\cite{sun2023pointnerfpp} & - & - & - & - & - & - & - & - & - \\
NeuRBF\cite{chen2023neurbf}* & 0.016 & 0.061 & \cellcolor{yellow!40}0.016 & 0.021 & \cellcolor{orange!40}0.015 & \cellcolor{orange!40}0.032 & 0.008 & 0.114 & 0.035 \\
\hline
RayGauss (ours) & \cellcolor{red!40}0.009 & \cellcolor{red!40}0.030 & \cellcolor{red!40}0.011 & \cellcolor{red!40}0.015 & \cellcolor{red!40}0.012 & \cellcolor{red!40}0.026 & \cellcolor{red!40}0.004 & \cellcolor{red!40}0.088 & \cellcolor{red!40}0.024 \\
  \end{tabular}
  }
  \caption{\textbf{PSNR, SSIM and LPIPS (with VGG network) scores on the Blender dataset.} All methods are trained on the train set with full-resolution images (800x800 pixels) and evaluated on the test set with full-resolution images (800x800 pixels).}
  \label{tab:detailed_blender}
\end{table*}

\begin{table*}[h]
  \centering
  \resizebox{0.8\linewidth}{!}{%
  \begin{tabular}{l | c c c c | c c c c c | c }
  & \multicolumn{9}{c}{\textbf{PSNR ↑}} \\
    & bonsai & counter & kitchen & room & bicycle & flowers & garden & stump & treehill & \textbf{Avg.}\\
    \hline
NeRF~\cite{Neural_Radiance_Fields}* & 22.10 & 22.34 & 22.00 & 24.46 & 19.35 & 19.49 & 22.70 & 21.43 & 21.02 & 21.65  \\
Instant-NGP~\cite{Instant_NGP}* & 27.04 & 24.25 & 23.44 & 27.30 & 23.69 & 21.41 & 25.64 & 22.56 & 22.22 & 24.17 \\
Gaussian Splatting~\cite{3D_Gaussian_Splatting}* & 33.42 & \cellcolor{orange!40}30.21 & \cellcolor{yellow!40}33.40 & \cellcolor{yellow!40}32.95 & \cellcolor{yellow!40}27.33 & \cellcolor{yellow!40}23.71 & 29.58 & \cellcolor{orange!40}27.78 & 24.00 & 29.15\\
Mip-Splatting~\cite{Yu2023MipSplatting}* & \cellcolor{yellow!40}33.44 & \cellcolor{red!40}30.43 & \cellcolor{orange!40}34.30 & \cellcolor{orange!40}33.30 & \cellcolor{orange!40}27.62 & \cellcolor{orange!40}23.79 & \cellcolor{yellow!40}29.78 & \cellcolor{red!40}27.89 & \cellcolor{yellow!40}24.25 & \cellcolor{yellow!40}29.42 \\
Zip-NeRF~\cite{zipnerf}* & \cellcolor{red!40}36.10 &    30.13 &    32.85 &    \cellcolor{red!40}34.20 &    \cellcolor{red!40}28.10 &    \cellcolor{red!40}24.25 &    \cellcolor{red!40}30.24 &    \cellcolor{orange!40}27.78 &    \cellcolor{red!40}25.72 & \cellcolor{red!40}29.93 \\
\hline
RayGauss (ours) & \cellcolor{orange!40}35.22 & \cellcolor{red!40}31.79 & \cellcolor{red!40}35.43 & \cellcolor{yellow!40}32.95 & 27.21 & 23.53 & \cellcolor{orange!40}29.91 & \cellcolor{yellow!40}27.13 & \cellcolor{orange!40}24.26 & \cellcolor{orange!40}29.71 \\
  \end{tabular}
  }

\bigskip

  \resizebox{0.8\linewidth}{!}{%
  \begin{tabular}{l | c c c c | c c c c c | c }
  & \multicolumn{9}{c}{\textbf{SSIM ↑}} \\
    & bonsai & counter & kitchen & room & bicycle & flowers & garden & stump & treehill & \textbf{Avg.}\\
    \hline
NeRF~\cite{Neural_Radiance_Fields}* & 0.652 & 0.690 & 0.658 & 0.815 & 0.371 & 0.462 & 0.653 & 0.482 & 0.506 & 0.588 \\
Instant-NGP~\cite{Instant_NGP}* & 0.923 & 0.769 & 0.736 & 0.920 & 0.658 & 0.604 & 0.829 & 0.563 & 0.611 & 0.735  \\
Gaussian Splatting~\cite{3D_Gaussian_Splatting}* & \cellcolor{yellow!40}0.970 & \cellcolor{yellow!40}0.942 & \cellcolor{orange!40}0.972 & 0.963 & 0.856 & 0.730 & \cellcolor{yellow!40}0.924 & \cellcolor{orange!40}0.833 & 0.734 & 0.880 \\
Mip-Splatting~\cite{Yu2023MipSplatting}* & \cellcolor{orange!40}0.971 & \cellcolor{orange!40}0.945 & \cellcolor{red!40}0.976 & \cellcolor{orange!40}0.966 & \cellcolor{red!40}0.871 & \cellcolor{orange!40}0.752 & \cellcolor{red!40}0.931 & \cellcolor{red!40}0.845 & \cellcolor{yellow!40}0.744 & \cellcolor{red!40}0.889 \\
Zip-NeRF~\cite{zipnerf}* & \cellcolor{red!40}0.978 & 0.932 & \cellcolor{yellow!40}0.951 & \cellcolor{yellow!40}0.965 & \cellcolor{orange!40}0.865 & \cellcolor{red!40}0.754 & 0.918 & \cellcolor{yellow!40}0.829 & \cellcolor{red!40}0.769 & \cellcolor{yellow!40}0.885 \\
\hline
RayGauss (ours) & \cellcolor{red!40}0.978 & \cellcolor{red!40}0.958 & \cellcolor{red!40}0.976 & \cellcolor{red!40}0.971 & \cellcolor{yellow!40}0.859 & \cellcolor{yellow!40}0.742 & \cellcolor{orange!40}0.929 & 0.810 & \cellcolor{orange!40}0.748 & \cellcolor{orange!40}0.886 \\
  \end{tabular}
  }

\bigskip
  
  \resizebox{0.8\linewidth}{!}{%
  \begin{tabular}{l | c c c c | c c c c c | c }
  & \multicolumn{9}{c}{\textbf{LPIPS ↓}} \\
    & bonsai & counter & kitchen & room & bicycle & flowers & garden & stump & treehill & \textbf{Avg.}\\
    \hline
NeRF~\cite{Neural_Radiance_Fields}* & 0.127 & 0.217 & 0.207 & 0.119 & 0.360 & 0.320 & 0.161 & 0.326 & 0.387 & 0.247 \\
Instant-NGP~\cite{Instant_NGP}* & 0.076 & 0.207 & 0.199 & 0.094 & 0.315 & 0.308 & 0.143 & 0.212 & 0.389 & 0.216 \\
Gaussian Splatting~\cite{3D_Gaussian_Splatting}* & 0.037 & 0.062 & \cellcolor{yellow!40}0.029 & 0.052 & 0.121 & 0.238 & \cellcolor{yellow!40}0.056 & \cellcolor{orange!40}0.142 & 0.230 & 0.107 \\
Mip-Splatting~\cite{Yu2023MipSplatting}* & \cellcolor{yellow!40}0.032 & \cellcolor{orange!40}0.057 & \cellcolor{orange!40}0.027 & \cellcolor{yellow!40}0.045 & \cellcolor{red!40}0.103 & \cellcolor{yellow!40}0.189 & \cellcolor{red!40}0.050 & \cellcolor{red!40}0.130 & \cellcolor{yellow!40}0.197 & \cellcolor{yellow!40}0.092 \\
Zip-NeRF~\cite{zipnerf}* & \cellcolor{red!40}0.021 & \cellcolor{yellow!40}0.059 & 0.034 & \cellcolor{orange!40}0.037 & \cellcolor{yellow!40}0.113 & \cellcolor{red!40}0.168 & 0.061 & \cellcolor{yellow!40}0.145 & \cellcolor{red!40}0.158 & \cellcolor{red!40}0.088 \\
\hline
RayGauss (ours) & \cellcolor{orange!40}0.024 & \cellcolor{red!40}0.042 & \cellcolor{red!40}0.024 & \cellcolor{red!40}0.036 & \cellcolor{orange!40}0.110 & \cellcolor{orange!40}0.183 & \cellcolor{orange!40}0.051 & 0.156 & \cellcolor{orange!40}0.187 & \cellcolor{orange!40}0.090 \\
  \end{tabular}
  }
  \caption{\textbf{PSNR, SSIM and LPIPS (with VGG network) scores on the Mip-NeRF 360 dataset.} All methods are trained and tested on downsampled images by a factor of 8.}
  \label{tab:detailed_mipnerf360}
\end{table*}


\end{document}